\documentclass[10pt]{article} %
\usepackage[accepted]{tmlr}

\usepackage{amsmath,amsfonts,bm}

\def\eqref#1{equation~\ref{#1}}

\def\1{\bm{1}}

\DeclareMathAlphabet{\mathsfit}{\encodingdefault}{\sfdefault}{m}{sl}
\SetMathAlphabet{\mathsfit}{bold}{\encodingdefault}{\sfdefault}{bx}{n}

\usepackage{hyperref}
\usepackage{url}

\usepackage[textsize=tiny]{todonotes}
\usepackage{float}
\usepackage{bm}
\usepackage[nameinlink]{cleveref}
\usepackage{fontawesome}
\usepackage{algorithm}
\usepackage{algpseudocode}
\usepackage{caption}
\usepackage{subcaption}

\usepackage{wrapfig}
\usepackage{tablefootnote}

\usepackage[utf8]{inputenc} %
\usepackage[T1]{fontenc}    %
\usepackage{booktabs}       %
\usepackage{amsfonts}       %
\usepackage{nicefrac}       %
\usepackage{microtype}      %
\usepackage{xcolor}         %

\usepackage{multirow}

\usepackage{listings}

\usepackage[normalem]{ulem}

\definecolor{myred}{RGB}{230, 0, 0}
\definecolor{mygreen}{RGB}{0, 229, 25}

\definecolor{codegreen}{rgb}{0,0.6,0}
\definecolor{codegray}{rgb}{0.5,0.5,0.5}
\definecolor{codepurple}{rgb}{0.58,0,0.82}
\definecolor{backcolour}{rgb}{0.95,0.95,0.92}

\lstdefinestyle{mystyle}{
    backgroundcolor=\color{backcolour},   
    commentstyle=\color{codegreen},
    keywordstyle=\color{magenta},
    numberstyle=\tiny\color{codegray},
    stringstyle=\color{codepurple},
    basicstyle=\ttfamily\footnotesize,
    breakatwhitespace=false,         
    breaklines=true,                 
    captionpos=b,                    
    keepspaces=true,                 
    numbers=left,                    
    numbersep=5pt,                  
    showspaces=false,                
    showstringspaces=false,
    showtabs=false,                  
    tabsize=2
}

\title{Towards Truly Zero-shot Compositional \\Visual Reasoning with LLMs as Programmers}

\author{\name Aleksandar Stani\'c\thanks{Work completed during an internship at Google Research.} \email alexstanic@google.com \\
      \addr Google DeepMind
      \AND
      \name Sergi Caelles  \email scaelles@google.com\\
      \addr Google DeepMind
      \AND
      \name Michael Tschannen \email tschannen@google.com \\
      \addr Google DeepMind
      }

\def\month{04}  %
\def\year{2024} %
\def\openreview{\url{https://openreview.net/forum?id=WYGiqSVstK}} %

\begin{document}

\maketitle
\begin{abstract}

Visual reasoning is dominated by end-to-end neural networks scaled to billions of model parameters and training examples.
However, even the largest models struggle with compositional reasoning, generalization, fine-grained spatial and temporal reasoning, and counting.
Visual reasoning with large language models (LLMs) as controllers can, in principle, address these limitations by decomposing the task and solving subtasks by orchestrating a set of (visual) tools.
Recently, these models achieved great performance on tasks such as compositional visual question answering, visual grounding, and video temporal reasoning.
Nevertheless, in their current form, these models heavily rely on human engineering of in-context examples in the prompt, which are often dataset- and task-specific and require significant labor by highly skilled programmers.
In this work, we present a framework that mitigates these issues by introducing spatially and temporally abstract routines and by leveraging a small number of labeled examples to automatically generate in-context examples, thereby avoiding human-created in-context examples.
On a number of visual reasoning tasks, we show that our framework leads to consistent gains in performance, makes LLMs as controllers setup more robust, and removes the need for human engineering of in-context examples.

\end{abstract}

\section{Introduction}

Compositional visual question answering requires a model to answer questions about visual content in a compositional manner, involving multiple steps of reasoning or considering relationships between different objects or entities within an image.
It is a complex task as it requires understanding both the visual information in an image and the structure of the question, and reasoning about how different visual elements relate to one another to generate the correct answer.
Recently, large progress has been made on many such vision and language tasks by scaling end-to-end neural networks models in terms of size, training data, and compute \citep{alayrac2022flamingo,chen2022pali,yu2022coca,wang2022git,gan2022vision,lu2022unified,li2023blip,driess2023palm,chen2023pali3,chen2023paliX}.
However, even the largest state-of-the-art (SotA) models struggle in tasks that require compositional reasoning, ability to generalize, fine-grained spatial reasoning capabilities, and counting \citep{bugliarello2023measuring,paiss2023teaching,hsieh2023sugarcrepe,yuksekgonul2022and,zhao2022vl,hendricks2021probing}.
An example of such task is the following query: ``Could the cookies on the table be equally distributed among children?'' \citep{suris2023vipergpt}.
To solve this, the model needs to detect the cookies in the image, filter out the ones that are not on the table, detect children, count cookies and children, and check if the cookies count is divisible by the children count.
Questions like these are difficult for current end-to-end vision and language models (VLMs).
Scaling VLMs further makes them even more data- and compute-hungry\citep{villalobos2022will}, so the scale alone seems unlikely to solve these tasks, especially due to the exponentially-large long tail of compositional tasks.

On the other hand, it is questionable whether solving compositional tasks with a single monolithic end-to-end neural network is the optimal approach.
Intuitively, it might be easier to first decompose the task into several subtasks, individually solve the subtasks, and then use the intermediate results to solve the original task.
This is reminiscent of the way humans approach compositional problems.
According to Daniel Kahneman's framework \citep{daniel2017thinking}, our thought process can be thought of as consisting of two mechanisms: System 1 and System 2.
System 2 is the ``slow'', ``analytical'' system that can decompose the task into subtasks, while System 1 is the ``fast'', ``reactive'' system that solves individual tasks such as recognizing patterns.
In machine learning, the early work on task decomposition was pioneered by Neural Module Networks (NMNs) \citep{andreas2016neural,johnson2017inferring,hu2017learning}.
NMNs are trained end-to-end in the hope that every module will learn a separate function that will be reusable across tasks.
However, these models have a number of drawbacks, namely that the program generation requires hand-tuned parsers, they are difficult to optimize (sometimes requiring reinforcement learning), and they have the issue of a module ``collapse'', where some modules are never activated and others take over all the work, contrary to the design intentions.

\begin{figure}[t!]
    \centering
    \begin{subfigure}[t]{0.47\textwidth}
        \centering
        \includegraphics[width=\textwidth]{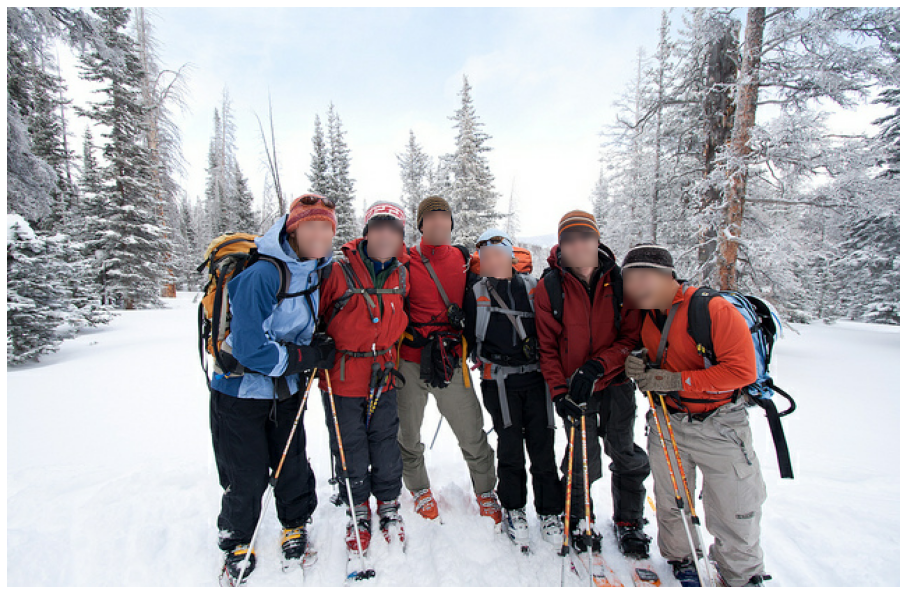}
        \caption{}
	\label{fig:refcoco_skiers}
    \end{subfigure}%
    ~
    \begin{subfigure}[t]{0.43\textwidth}
        \centering
        \includegraphics[width=\textwidth]{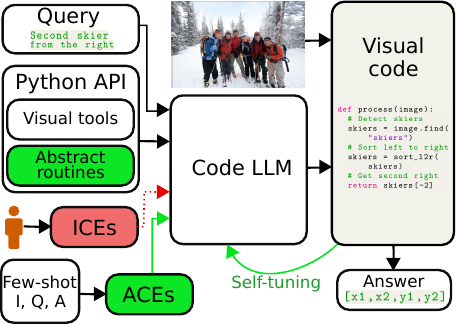}
        \caption{}
	\label{fig:model_diagram}
    \end{subfigure}
    \caption{
    (a) RefCOCO \citep{yu2016modeling} example image.
    (b) A code-generating LLM takes as input the query, the Python API (functions for ``tool use'' and \textcolor{mygreen}{Abstract API routines (functions)} we introduce in \Cref{sec:abstract_api}) and a number of \textcolor{myred}{ICEs} (we replace human-engineered \textcolor{myred}{ICEs} by automatically-generated \textcolor{mygreen}{ACEs} in \Cref{sec:method_aice}).
    The LLM generates code that takes as input the image and outputs an answer (here a bounding box).
    If code fails to run, \textcolor{mygreen}{``self-tuning''} (\Cref{sec:self-correction}) can adjust parameters and generate new code.
    }
    \label{fig:main_figure}
\end{figure}

Recently, an alternative approach based on ``tool use'' gained popularity \citep{cobbe2021training,komeili2021internet,thoppilan2022lamda,parisi2022talm,zeng2022socratic,gao2023pal,qin2023toolllm,zhuge2023mindstorms}.
In ``tool use'', an LLM solves a task by controlling (akin to System 2) a set of tools (such as an object detector, akin to System 1) \citep{zeng2022socratic,shen2023hugginggpt,zhuge2023mindstorms}.
In particular, VisProg \citep{gupta2023visual}, ViperGPT \citep{suris2023vipergpt}, and CodeVQA \citep{subramanian2023modular} show great promise in solving visual question answering by using an LLM to generate a program (in Python or a custom scripting language).
During execution, the program calls individual vision models (such as object detector, depth estimator) through an \textit{API} that is provided in the prompt.
For example, to answer ``Which color is the jacket of the second person from the left?'' (\Cref{fig:refcoco_skiers}), the program needs to detect people, sort them from left to right, select the second, detect their jacket, and query its color.
These models achieved SotA on compositional visual question answering, visual grounding, and video temporal reasoning tasks.
By their construction, they are interpretable, compositional, adaptable (tools can be upgraded on the fly), offer strong generalization, mathematical, and reasoning skills, and do not require gradient-based training.
However, in their current form, they heavily rely on human engineering of in-context (program) examples (ICEs) in the prompt.
Moreover, ICEs are dataset- and task-specific.
To generate them, significant labor by highly skilled programmers is required.
For this reason, we argue that these methods \emph{should not be called zero-shot} in their current form.

In this work, we present a framework that mitigates these issues, makes LLMs-as-programmers setup more robust, and removes the need for human engineering of ICEs.
Our framework, whose effectiveness we show across a number of compositional question-answering and video temporal reasoning tasks with ViperGPT (but is universally applicable to other approaches), consists of the following:
\begin{itemize}
    \item 
    Firstly, instead of using a simple API with only basic routines that call individual tools, we introduce an ``Abstract API''.
    Abstract API consists of spatially and temporally abstract routines that remove the large burden on the LLM to have strong spatial and temporal reasoning.
    \item 
    Second, instead of relying on a large number of dataset-specific (question, code)-pairs as ICEs, we introduce a setup that generates ICEs automatically.
    Using a few labeled examples (that are significantly cheaper to obtain, e.g. via crowd-sourcing), we generate query-code examples in a \emph{zero-shot} manner and use these as ICEs.
    This mitigates the need for human engineering of ICEs.
    \item 
    Third, we show how LLMs as controllers for visual reasoning can (to some extent) perform ``self-correction'' through ``self-debugging'' and ``self-tuning'' without any ground truth labels.
    In ``self-debugging'', we generate new code when the previous fails to execute, either by providing the LLM previous query-code pair and execution error as feedback or from scratch.
    In ``self-tuning'', we show how the tool hyperparameters can be tuned automatically if the execution fails due to a module.
\end{itemize}

\section{LLMs as programmers for visual reasoning framework}

In this section, we first provide a brief overview of the ViperGPT approach \citep{suris2023vipergpt} on top of which we show the utility of our framework.
We then describe each component of our framework, namely the Abstract API, automatic generation of ICEs, and self-correction.

\subsection{Background}
\label{sec:background}

ViperGPT takes as input an image or a video and a textual query.
The textual query is fed into an LLM (Codex \citep{chen2021evaluating}), together with the tools API and ICEs.
The LLM generates a program that solves the given query using the tools without further training.
The information in the prompt is crucial for good ViperGPT performance, as it is the only task-specific information provided.
The prompt consists of a Python API with the necessary functions to solve the visual query, such as object detection, depth estimation, and language model queries.
Additionally, ViperGPT uses several dataset-specific ICEs in the prompt.
As we show in \Cref{sec:experiments}, performance depends heavily on these human-engineered examples.

ViperGPT API defines an \texttt{ImagePatch} and a \texttt{VideoSegment} class that contain image and video processing functions.
Each function calls a pretrained model to compute the result.
The API in the prompt does not contain function implementations, but it contains docstrings and query-code examples of their use.
The ViperGPT API defines the following functions:
\texttt{find} takes as input an image and a textual query, calls an open vocabulary detector and returns a list of image patches of detected objects;
\texttt{exists} takes as input an image and a textual query and returns true if the query object exists in the image, otherwise false; 
\texttt{verify\_property} takes as input an image, a noun representing an object and an attribute property to verify and returns a boolean whether the object has this property;
\texttt{best\_image\_match} that takes as input a list of image patches and a textual query and returns the image patch that best matches the query;
\texttt{best\_text\_match} that takes as input a list of queries and one image, and returns the query that best matches the image;
\texttt{compute\_depth} that computes the median depth of an image or image patch;
\texttt{distance} which computes the pixel-distance between two images;
\texttt{simple\_query} which handles textual queries that are not decomposable, by calling an image captioning model;
\texttt{select\_answer} that given a context text describing a scene and a list of possible answers queries an LLM to select the correct answer.
The \texttt{VideoSegment} class does not contain any functions that call individual models, but only the start and end point of the video segment and an iterator over the frames, which returns an \texttt{ImagePatch} object.
For the full ViperGPT API, see \Cref{sec:prompt_listings}.

The code-generating LLM outputs code that attempts to solve the query.
This code is executed, taking as input an image or a video (and optionally a list of possible answers) and outputs a result (e.g. a bounding box or a string).
Due to generating programs in native Python code, ViperGPT avoids the need for custom interpreters and can leverage Python built-in functions (e.g. sort, if/else control flows, math functions, etc.).

\subsection{Abstract API through visual routines}
\label{sec:abstract_api}

When programming, we continuously build new layers of abstraction.
grouping them together into new functions.
By abstracting away the implementation details, we reduce the cognitive load on the programmer which is then able to build systems of increased complexity.
Motivated by this, we introduce a set of \emph{spatially} and \emph{temporally} abstract functions 
(routines \footnote{Note that our ``routines'' do not correspond to the visual routines of \cite{ullman1987visual} such as tracing or scanning.})
that abstract away a number of lines of code for the same functionality and together make the \emph{Abstract API}.
From a practical perspective, we are motivated by a number of failure cases observed in the experiments (see \Cref{sec:experiments}).
As presented in \Cref{sec:background}, ViperGPT's API is fairly simple (contains almost exclusively functions to call pretrained models).
Although simplicity is good and often desirable, in the case of visual reasoning with LLMs as programmers, the lack of expressive visual routines requires the code-generating LLM to have strong spatial and temporal reasoning capabilities.
Qualitative analysis showed that this is often not the case and that the current LLMs are not yet perfect in these terms (e.g. they confuse left and right, top and bottom).
For example, for the query ``return the second person from the right'', the program generated by the LLM correctly sorts the persons along the horizontal axis but then wrongly takes the second index in the array (instead of the second last).
Similarly, they sometimes ``confuse'' temporal order, e.g., if a ``before'' event means a smaller or a larger time index.

For these reasons, we introduce a set of spatially and temporally abstract routines.
We add the following spatial routines:
\texttt{get\_patch\_left\_of},
\texttt{get\_patch\_right\_of},
\texttt{get\_patch\_above\_of},
\texttt{get\_patch\_below\_of}
for relational retrieval relative to a patch;
\texttt{get\_patch\_closest\_to\_anchor\_object} that sorts patches by their distance to an anchor object and returns the one with the smallest distance;
\texttt{sort\_patches\_left\_to\_right},
\texttt{sort\_patches\_bottom\_to\_top}, and
\texttt{sort\_patches\_front\_to\_back} to sort the list of patches along horizontal, vertical or depth axis;
\texttt{get\_middle\_patch} to get the middle patch from a given list of image patches;
For videos, we add temporal routines for event ``localization'':
\texttt{get\_video\_segment\_of\_event}, 
\texttt{get\_video\_segment\_before\_event},
\texttt{get\_video\_segment\_after\_event}, and routines to either caption a video: 
\texttt{caption\_video} or answer a simple question about the video: 
\texttt{simple\_query}.
The routines that we introduce are \emph{general} in the sense that they are not specific to any individual task or dataset.
This facilitates their reuse across tasks and avoids engineering task and dataset-specific routines.
It is an open research question what the ``optimal'' set of primitive routines is.
Another exciting research direction is using LLM with their own abstract routines, then reusing those to come up with even more abstract routines and so on.
We leave these for future work.

\subsection{Automatic generation of in-context examples}
\label{sec:method_aice}

In-context examples (ICEs) greatly influence the performance of LLMs \citep{brown2020language, chen2023demonstrations}.
For example, ViperGPT \citep{suris2023vipergpt} uses between 10 and 16 hand-engineered dataset-specific query-code ICEs per dataset.
Similarly, VisProg \citep{gupta2023visual} uses between 16 and 31 ICEs and CodeVQA \citep{subramanian2023modular} about 50 ICEs.
However, constructing these ICEs requires heavy human engineering, as they might need to be rewritten in a way that the LLM can use them to ``generalize'' to other examples across the dataset.
Furthermore, the constructed examples are specific not only to the dataset but also to the LLM and the API.
If any of those changes, they need to be written from scratch.
Finally, to write good query-code ICEs, highly skilled labor is required, ideally someone familiar with the workings of LLMs and a good grasp of Python programming.

In our work, we move beyond this need for human engineering of query-code ICEs.
We start from a small set of labeled examples (e.g. 16 image-question-answer tuples), as is common in few-shot transfer learning \citep{zhai2019visual,kolesnikov2020big}.
We run our framework in a \emph{zero-shot} manner (without any ICEs) on these few-shot examples, sort the results by accuracy, select the best-performing programs, pair them with the corresponding queries, and use them as ICEs during test time.
We call such ICEs \emph{automatically-generated in-context examples} (ACEs).
Importantly, \emph{no gradient-based optimization is performed on the few-shot examples}.
Intuitively, this works since even if the LLM does not always generate a program that solves the task correctly, it might sometimes come up with a correct program.
Since retrieval is often easier than generating programs from scratch, the reuse of the correct programs improves performance on the test set.

ACEs provide a number of benefits over manually writing ICEs.
First of all, ACEs are much cheaper to obtain as they do not require highly skilled labor to write them.
Second, the algorithm that generates ACEs is general: it is neither specific to the API nor the LLM.
If any of these changes, ACEs can be easily generated by re-running the algorithm.
Furthermore, they can be seen as a first step of the LLM ``coming up'' with its own abstract rules and thus creating a ``rulebook'' (discussed in \Cref{sec:abstract_api}).
Finally, few-shot (image, question, answer)-labeled examples are often available in datasets typically used in machine learning.
If not available, they are cheap to obtain via crowd-sourcing and can be reused for further studies as a benchmark.

\subsection{Self-correction}
\label{sec:self-correction}

One of the advantages of solving visual reasoning tasks with LLMs as programmers is that we know when code fails to execute.
The failure can happen, e.g. due to a compilation error (e.g. due to hallucination), some of the models failing, or a wrong return type (e.g. a bounding-box is expected, but code returns a string).
Note that to detect these types of errors, no ground-truth labels are needed.

\paragraph{Self-debugging.}
If the code execution fails, we can query the code-generating LLM to correct the previously generated code.
We do this by feeding back the query, the previously generated code, and the resulting error in the prompt (see the feedback template in \Cref{sec:self-debugging-prompt}).
Moreover, if the LLM's sampling temperature is higher than zero, we can query the model with a different random seed to generate new code from scratch.
There are advantages to both of these approaches.
If the code-generating LLM has good ``self-correction'' abilities, then it should be able to correct its own mistakes based on the feedback, as we humans could.
However, if the LLM is not good at self-correction or does not know how to incorporate such feedback (e.g. if the LLM is not trained to be conversational), then feeding back the previous query and code will only ``bias'' the model to output the same solution.
In that case, generating new code from scratch could work better.

\paragraph{Self-tuning.}
In some cases, we know that code execution failed due to some components of a particular module.
For example, the open vocabulary detector fails due to a too high threshold hyperparameter.
When the threshold is high, we have a higher number of false negatives.
For such cases, we propose to automatically change the hyperparameter of the module (e.g. reduce the threshold) and execute code again.
Although here we experiment with only a single option of tuning such hyperparameters, the idea is applicable to any model that involves such ``sensitivity'' hyperparameters.
For example, another such candidate could be the threshold that determines the ``similarity'' of the CLIP-style image-text embedding model.
The task of this module is to filter out a particular instance (e.g. ``blue ball'') from a set of instances (``balls''), e.g., based on an attribute (``blue'').
It does so by comparing the text embedding of ``blue'' with the image embeddings of all ``ball'' image patches in the image.
Then it returns all image patches whose embeddings have similarity with the text embedding higher than the threshold value.
Here we would similarly tune the threshold in case this module fails.

\section{Experiments}
\label{sec:experiments}

\paragraph{Tasks.}

We evaluate our method on four datasets: RefCOCO, RefCOCO+ \citep{yu2016modeling}, GQA \citep{hudson2019gqa} and NExT-QA \citep{xiao2021next} used in previous work \citep{suris2023vipergpt}.
These datasets evaluate a diverse set of capabilities, namely visual grounding (RefCOCO, RefCOCO+), compositional image question answering (GQA), and video temporal reasoning (NExT-QA).
In RefCOCO (example in \Cref{fig:refcoco_skiers}), the task is to detect a bounding box of an object given its natural language description (``referring expression'').
In compositional question answering in GQA, the task is to answer in natural language a compositional natural language query.
We use the ``test-dev'' split of the GQA dataset, as in ViperGPT.
In NExT-QA, the task is to answer a temporally compositional question by selecting one of the given multiple choice options.
As in ViperGPT, we use NExT-QA ``hard'' split \cite{buch2022revisiting}.
For RefCOCO and RefCOCO+, methods are evaluated in terms of Intersection over Union (IoU) between the detected and the ground truth bounding box and for GQA and NExT-QA in terms of accuracy of the predicted answer.

\paragraph{Vision and Language Models.}

For code generation, we use a code instruction-tuned version of PaLM 2 \citep{anil2023palm} \texttt{code-bison} accessible via the Google Cloud API \citep{google2023codebison}.
We use the same model to select an answer for the multichoice questions in the NExT-QA dataset.
Vision models we use are OWLv2 \citep{minderer2023scaling} for object detection, SigLiT \citep{zhai2023sigmoid} for text-image comparison, MiDaS \citep{ranftl2020towards} for depth estimation, and PaLI-3 \citep{chen2023pali3} for image captioning and answering visual queries.
Note that all models are different from the models that ViperGPT used (see \Cref{sec:pretrained_models}).

\begin{table}[t]
\caption{
Comparison of our method against previous end-to-end and ``LLMs as controllers'' SotA methods.
For ``Ours (\texttt{code-bison})'', we report mean scores $\pm$ standard deviation across three random seeds.
The reference numbers for SotA on each dataset are taken from the following publications:
RefCOCO: ZS \citep{yang2023fine}, FS \citep{yao2021cpt}, Sup \citep{wang2022ofa};
RefCOCO+: ZS \citep{yang2023fine}, FS \citep{yao2021cpt}, Sup \citep{wang2022ofa};
GQA: ZS \citep{li2023blip}, FS \citep{jin2021good}, Sup \citep{nguyen2022coarse};
NExT-QA: ZS \citep{chen2023pali3}, FS \citep{chen2023pali3}, Sup \citep{ye2023hitea}.
}
\label{tbl:baselines}
\begin{center}
\begin{tabular}{lcccc}
\toprule
Model                        & RefCOCO (IoU) & RefCOCO+ (IoU) & GQA (acc.) & NExT-QA (acc.)\\
\midrule
Zero-Shot (ZS) SotA          & 53.0          & 57.5           & 44.7       & 38.3 \\
Few-Shot (FS) SotA                & 53.3          & 52.5           & 35.7       & 38.3 \\
\textcolor{gray}{Supervised (Sup) SotA}  & \textcolor{gray}{94.0}          & \textcolor{gray}{91.7}           & \textcolor{gray}{72.1}       & \textcolor{gray}{63.1} \\
\midrule
ViperGPT (paper)             & 72.0          & 67.0           & 48.1       & 60.0 \\
ViperGPT (GitHub (GH) ZS)    & 46.7          & -              & -          & -    \\
\textcolor{gray}{ViperGPT (GH w/ DS-ICEs)}  & \textcolor{gray}{60.5}          & -              & -          & -    \\
\midrule
E2E bsl.(ZS OWLv2/PaLI-3)    & 33.5          & 31.7           & 40.1       & 58.9 \\
E2E LLM-only baseline        & -             & -              & -          & 53.3 \\
\midrule
Ours (\texttt{code-bison}, Zero-Shot) & 44.4 {\scriptsize $\pm$ 0.9} & 38.2 {\scriptsize $\pm$ 0.0} & 32.1 {\scriptsize $\pm$ 0.4} & 60.2 {\scriptsize $\pm$ 0.3} \\
Ours (\texttt{code-bison})            & 51.2 {\scriptsize $\pm$ 0.2} & 45.7 {\scriptsize $\pm$ 0.1} & 33.4 {\scriptsize $\pm$ 0.2} & 61.0 {\scriptsize $\pm$ 0.1} \\
\bottomrule
\end{tabular}
\end{center}
\end{table}

\paragraph{Baselines.}
Strong baselines are essential for correctly measuring progress in machine learning.
This is especially true in the emerging area of ``tool use'' \citep{cobbe2021training,komeili2021internet,thoppilan2022lamda,parisi2022talm,zeng2022socratic,gao2023pal,qin2023toolllm,zhuge2023mindstorms}.
When using an LLM and other pre-trained models, we must be careful to report the exact LLM version and/or API when it was accessed, and ideally report results over several random seeds to measure the statistical significance.
In \Cref{tbl:baselines}, we provide an overview of the previous Zero-Shot (ZS), Few-Shot (FS), and Supervised (Sup) SotA methods, ViperGPT, end-to-end (E2E) baselines, and our results on all datasets we used for evaluation.

Early in the project, we found it difficult to reproduce the results reported in the ViperGPT paper.
Our first hypothesis is that this is due to differences in the vision and language models we use compared to the ViperGPT paper.
However, when running the original ViperGPT code from the official GitHub repository on RefCOCO, we were only able to achieve an IoU of 60.5 as opposed to 72.0 reported in the paper.
Note, however, that ViperGPT uses Codex, which is discontinued, so we use \texttt{GPT-3.5-turbo} \citep{openai2023gpt35}.
Also note that this score was obtained using 16 \textit{dataset-specific} ICEs (DS-ICEs).
These examples contain large amounts of dataset-specific human-engineered information, such as ``clothing requires returning the person''.
In the case of truly Zero-Shot learning (without any human-engineered ICEs), the IoU score of ViperGPT's official GitHub code drops by 14 points to 46.7.
Moreover, in their GitHub code we found hand-engineered improvements:
if returning a bounding box fails, then return an ``average'' bounding box, and if code execution fails on GQA, then query the image captioner (BLIP-2 \citep{li2023blip}).
These code changes lead to improved results, but make it hard to quantify the true power of LLMs as controllers approach.

In \Cref{tbl:baselines}, we also provide Zero-Shot end-to-end baselines.
On RefCOCO and RefCOCO+, we feed the query as input to the OWLv2 model and return its output.
For the GQA end-to-end baseline, we query the PaLI-3 model (which was fine-tuned for multi-task inference on different captioning and VQA data sets, but not on GQA).
For NExT-QA, we provide two baselines.
For the first baseline, we subsample and caption with PaLI-3 one frame per second, and then feed all these captions to an LLM (\texttt{code-bison}) together with the question and multiple choice answers.
As the second baseline, we simply feed the LLM the question and the possible answers and ask it to choose one.
LLM-only baseline achieves an accuracy of 53.3\%, which is only 5.4\% lower than the baseline that also gets videos as input and significantly above the chance level accuracy of 20\% (since there are 5 possible multichoice answers to each question).
This suggests that NExT-QA might not fully evaluate the vision properties of the model and that new datasets are needed; for example, the Perception Test \citep{puatruaucean2023perception} was created specifically to avoid such problems.

Lastly, in \Cref{tbl:baselines} we report the Zero-Shot performance in our setup, as well as results for our best performing model variant (averaged over three random seeds).
In the following sections, we evaluate each component, as well as their combinations, to obtain the reported results.
Using \texttt{GPT-3.5-turbo} instead of \texttt{code-bison} resulted in a slight drop in performance, but as we shall see below, the same conclusions hold for both \texttt{code-bison} and \texttt{GPT-3.5-turbo} for all our suggested improvements.
In the following, we present the components of our framework.
For an ablation study, see \Cref{tbl:ablation-componentwise} that shows that each component contributes positively to the final scores on each dataset.

\subsection{Zero-Shot through spatially and temporally Abstract API}

\begin{figure}[t!]
    \centering
    \begin{subfigure}[t]{0.24\textwidth}
        \centering
        \includegraphics[width=\textwidth]{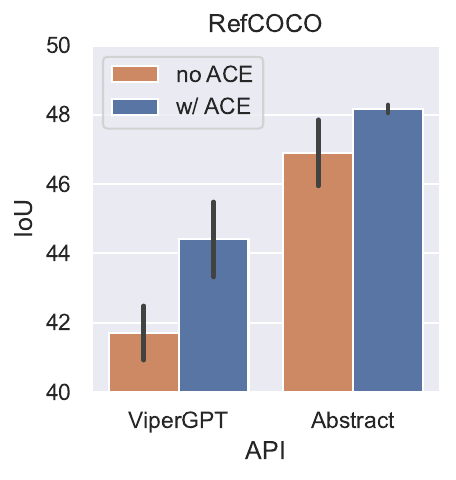}
    \end{subfigure}%
    ~
    \begin{subfigure}[t]{0.24\textwidth}
        \centering
        \includegraphics[width=\textwidth]{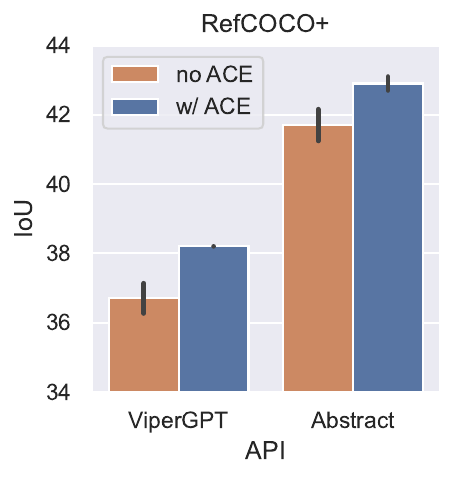}
    \end{subfigure}%
    ~
    \begin{subfigure}[t]{0.24\textwidth}
        \centering
        \includegraphics[width=\textwidth]{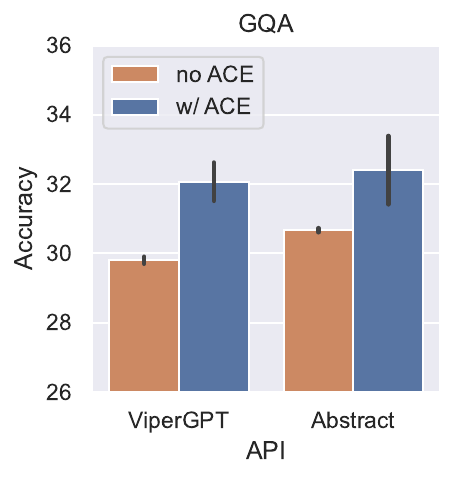}
    \end{subfigure}%
    ~
    \begin{subfigure}[t]{0.24\textwidth}
        \centering
        \includegraphics[width=\textwidth]{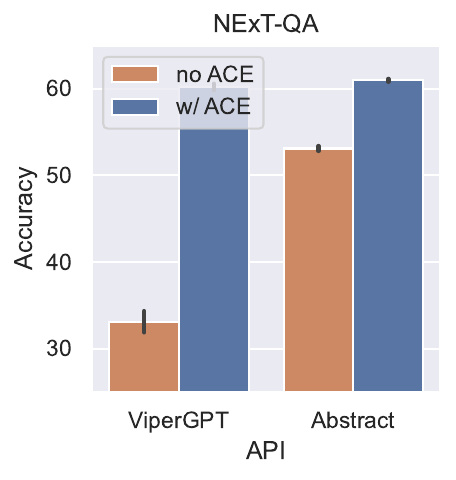}
    \end{subfigure}
    \caption{
    Using our Abstract API improves performance over the ViperGPT API across all datasets.
    Similarly, ACEs consistently improve performance, and these gains compound with the gains from the Abstract API.
    Uncertainty bars represent standard deviations computed over three random seeds.
    }
    \label{fig:api_and_aice}
\end{figure}

In \Cref{fig:api_and_aice}, we show the effect of using our Abstract API instead of the API used in ViperGPT.
The API for RefCOCO, RefCOCO+, and GQA uses the same set of image routines (see \Cref{sec:prompt_listings}), whereas the API for NExT-QA uses only video-specific (temporal) routines.
For now, we focus on the brown bars in \Cref{fig:api_and_aice}, and we compare the ViperGPT API and our Abstract API.
We can see that our Abstract API leads to gains both with and without ACEs across all datasets.
The performance gain is most notable for the NExT-QA dataset when ACEs are not used.
We suspect that this is due to the LLM's difficulty in reasoning about the temporal order of events.
This confirms our hypothesis that building a more abstract API such that the LLM does not need to use low-level Python routines is a promising direction.

Finally, we investigate whether our conclusions also hold for other LLMs, namely OpenAI's \texttt{GPT-3.5-turbo}.
On RefCOCO \texttt{GPT-3.5-turbo} achieves an IoU of 28.9 with the ViperGPT API and 39.8 with our Abstract API and an accuracy of 9.4 and  42.9 for the ViperGPT API and our Abstract API, respectively, on NExT-QA.
This confirms that our Abstract API brings gains not only for \texttt{code-bison}, but also for other LLMs.
For each sample in the evaluation, we allow only one trial of code generation (no self-correction).
Compared to the results with \texttt{code-bison}, IoU of 37.7 and 40.5 on RefCOCO and accuracy of 11.5 and 46.1 on NExT-QA for the ViperGPT API and our Abstract API, respectively, the results with \texttt{GPT-3.5-turbo} are slightly worse.
We suspect that the reason for this could be that \texttt{GPT-3.5-turbo} is mainly built to be a conversational agent, while \texttt{code-bison} is specifically trained to have good coding capabilities.

\subsection{Few-shot boostrapping via automatically generated in-context examples (ACEs)}

We evaluate the effect of using automatically generated in-context examples (ACEs), described in \Cref{sec:method_aice}.
We can either sample few-shot examples manually or pick them at random.
Both of these variants lead to good results, as we show in the following experiments.
However, selecting examples manually allows for a better ``quality'' (in terms of diversity) given a small number of few-shot examples, so by default we use these for all experiments.
For the first set of experiments, we manually pick 16 few-shot examples from the training set: image/video, question, and ground-truth answer.
We try to make examples diverse to cover question ``types'' (e.g. left-right, front-back, closest-to, etc.).

\begin{figure}[t]
    \centering
    \begin{subfigure}[t]{0.24\textwidth}
        \centering
        \includegraphics[width=\textwidth]{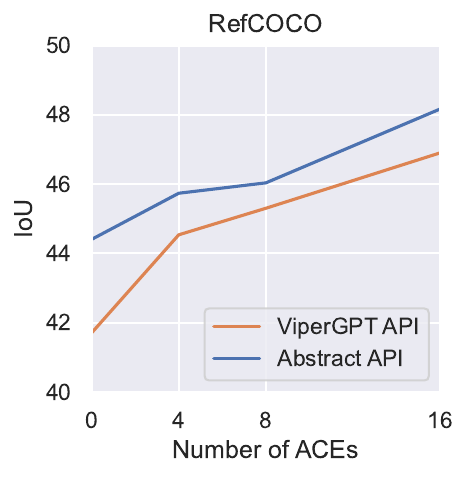}
    \end{subfigure}%
    ~
    \begin{subfigure}[t]{0.24\textwidth}
        \centering
        \includegraphics[width=\textwidth]{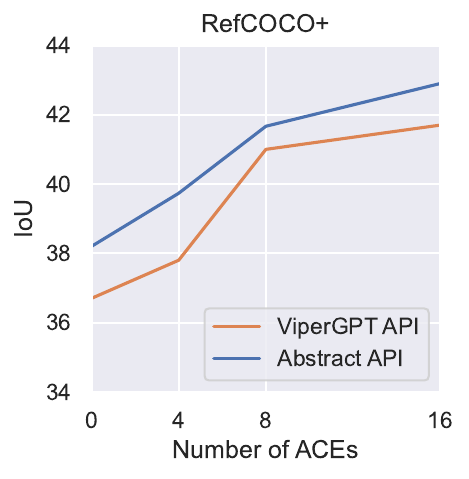}
    \end{subfigure}%
    ~
    \begin{subfigure}[t]{0.25\textwidth}
        \centering
        \includegraphics[width=\textwidth]{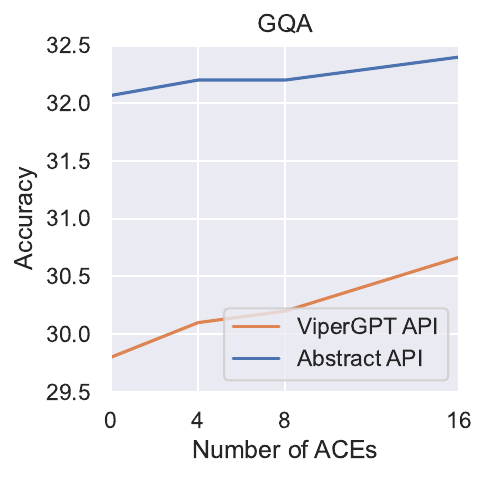}
    \end{subfigure}%
    ~
    \begin{subfigure}[t]{0.24\textwidth}
        \centering
        \includegraphics[width=\textwidth]{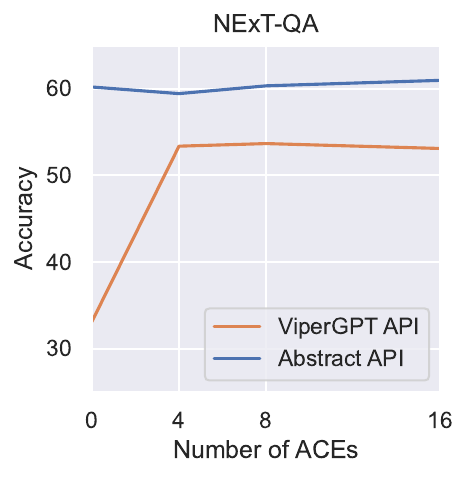}
    \end{subfigure}
    \caption{
    Increasing the number of ACEs in the prompt improves performance.
    Note that using the ViperGPT API on NExT-QA results in only three correct ACEs, so the performance plateaus after four ACEs.
    }
    \label{fig:aice04816}
\end{figure}

In \Cref{fig:api_and_aice}, we show the effect of ACEs.
For each dataset and for each API, the performance without ACEs is shown with the brown bar, and the performance with ACEs corresponds to the blue bar.
We can see that for all datasets and all APIs, ACEs improve performance.
The largest gains when using ACEs are for RefCOCO, RefCOCO+, and NExT-QA datasets when using the ViperGPT API.
This indicates that ACEs are effective in dealing with complex spatial and temporal reasoning.
More importantly, it can be seen in \Cref{fig:api_and_aice} that the gains from both the Abstract API and ACEs compound for all tasks, indicating that they provide complementary strengths.
\Cref{fig:aice04816} shows how the performance in terms of IoU and accuracy scales with the number of few-shot examples used to generate ACEs.
As expected, increasing the number of few-shot examples leads to improved performance.
Note that the ViperGPT API on NExT-QA is able to correctly ``solve'' only 3 few-shot examples, so there are no gains beyond using 4 few-shot examples.

We evaluate the effect of using randomly sampled few-shot examples instead of manually selecting them.
On RefCOCO, we sample 100 few-shot random samples from the training set, run Zero-Shot framework on them, sort the resulting programs by their IoU, and select the top 16 programs.
Therefore, we end up with the same number of ACEs as with manual selection.
On RefCOCO, we achieve IoU of 47.9 and 49.1 with the ViperGPT API and our Abstract API respectively.
These results are slightly better than those with manually selected few-shot examples (46.9 and 48.2 IoU).
This shows that the manual labor for generating ACEs can be removed altogether if we already have some labeled examples.
With 50 few-shot random samples we obtain similar performance, and for 16 such samples we observe a small drop in performance (see \Cref{tbl:random-few-shot-samples-refcoco,tbl:random-few-shot-samples-refcoco-plus} in the \Cref{sec:ablations} for detailed results).

As in the previous section, we test whether our findings are consistent with \texttt{GPT-3.5-turbo} on RefCOCO and NExT-QA.
On RefCOCO, when using \texttt{GPT-3.5-turbo}, ACEs improve IoU from 28.9 to 39.8 with the ViperGPT API and from 38.1 to 41.6 with our Abstract API.
Similarly, for \texttt{GPT-3.5-turbo} on NExT-QA, ACEs improve accuracy from 9.4 to 42.9 the ViperGPT API and from 56.7 to 58.8 with our Abstract API.
This confirms that the benefit of ACEs is not only limited to \texttt{code-bison} but also holds for \texttt{GPT-3.5-turbo} as well.

Another benefit of the few-shot setup when generating ACE is that it allows us to ``tune'' hyperparameters (HPs).
For example, when sweeping over LLM temperature and object detection threshold HPs, we observed that the relative performances on the few-shot examples closely resemble the one when sweeping over the full validation dataset (the analysis is shown in \Cref{sec:ablations}).

\subsection{Self-correction}

In this section, we analyze the ability of the framework to ``self-correct'' itself \emph{without any external feedback} when code execution fails.

\begin{figure}[t!]
    \centering
    \begin{subfigure}[t]{0.32\textwidth}
        \centering
        \includegraphics[width=\textwidth]{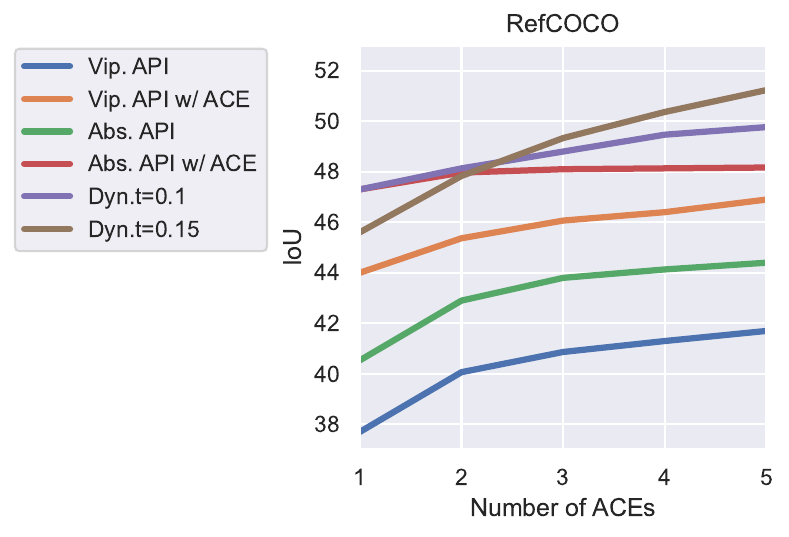}
    \end{subfigure}%
    ~
    \begin{subfigure}[t]{0.21\textwidth}
        \centering
        \includegraphics[width=\textwidth]{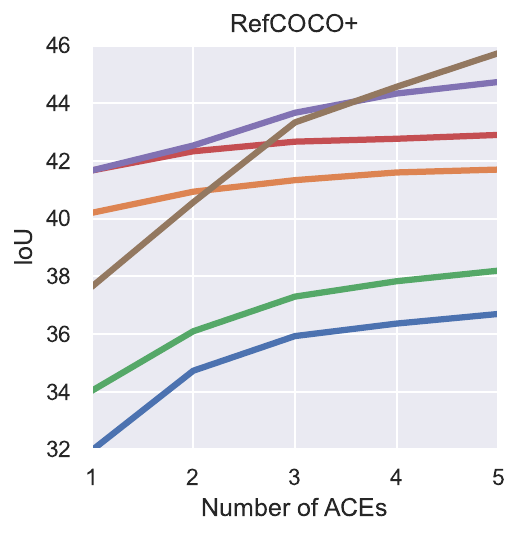}
    \end{subfigure}%
    ~
    \begin{subfigure}[t]{0.21\textwidth}
        \centering
        \includegraphics[width=\textwidth]{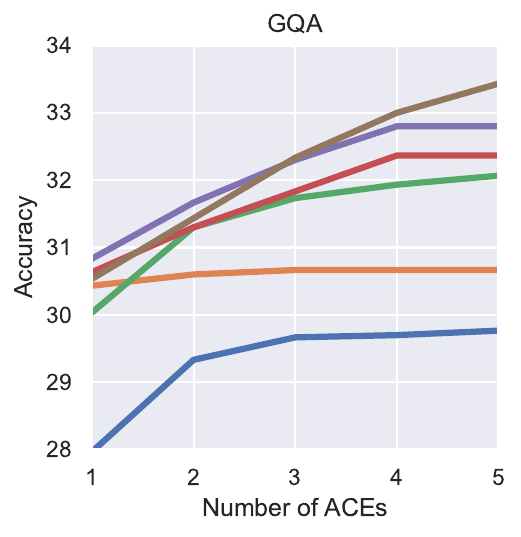}
    \end{subfigure}%
    ~
    \begin{subfigure}[t]{0.21\textwidth}
        \centering
        \includegraphics[width=\textwidth]{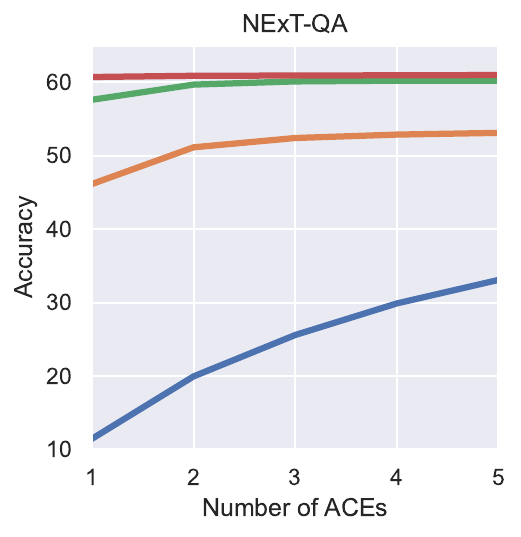}
    \end{subfigure}
    \caption{
    Increasing the number of ``self-tuning'' steps leads to improved performance.
    Our Abstract API (Abs. API) consistently outperforms the ViperGPT API (Vip. API).
    The best performance is achieved when using dynamic object detector threshold (Dyn.t) in addition to the Abstract API with ACE.
    }
    \label{fig:selftune}
\end{figure}

\begin{table}[t]
\caption{
Component-wise ablations of our framework.
Each component contributes positively to the final score.
Their relative contributions vary for different tasks.
We report mean scores across three random seeds.
}
\label{tbl:ablation-componentwise}
\begin{center}
\begin{tabular}{lllll}
\toprule
Model            & RefCOCO (IoU) & RefCOCO+ (IoU) & GQA (acc.)  & NExT-QA (acc.) \\
\midrule
ViperGPT API     & 38.4          & 32.0           & 27.9        & 11.5           \\
+ Abstract API   & 42.3 (+3.9)   & 34.0 (+2.0)    & 30.0 (+2.1) & 57.6 (+46.1)   \\
+ ACE            & 47.3 (+5.0)   & 41.7 (+7.7)    & 30.6 (+0.6) & 60.7 (+3.1)    \\
+ Self-debugging & 48.2 (+0.9)   & 42.9 (+1.2)    & 32.4 (+1.8) & \textbf{61.0} (+0.3)    \\
+ Self-tuning    & \textbf{51.2} (+3.0)   & \textbf{45.7} (+2.8)    & \textbf{33.4} (+1.0) & -              \\
\bottomrule
\end{tabular}
\end{center}
\end{table}

\paragraph{Self-debugging.}
When the program execution fails (due to e.g. compilation errors), we can retry by generating a new program \citep{chen2021evaluating}.
When creating a new program, we can also feed the previously generated code and the question as part of the prompt (see \Cref{sec:prompt_listings}), a variant that we call ``self-debugging''.
Another option would be to simply repeat the exact same prompt as in the previous trial and rely on stochasticity in the LLM with a temperature greater than zero to generate a new correct solution.
In our experiments, the ``self-debugging'' variant did not lead to an improvement in performance.
In all cases, the performance plateaus after the first trial.
This is in line with other recent findings \cite{huang2023large,stechly2023gpt4,valmeekam2023large}.
On the other hand, the variant without any feedback in the prompt led to an increasingly better performance as the number of ``trials'' increases (see \Cref{fig:selftune}).

\paragraph{Self-tuning.}
In some cases, we know that code fails due to some specific module.
Therefore, we can then adjust the hyperparameters of the respective module and re-run code.
For example, the open vocabulary detector we used (OWLv2) has a threshold hyperparameter that controls how sensitive it is.
The lower this threshold, the more false positives we will have, but also the fewer false negatives.
There is no global ``optimal'' value for this threshold that performs best for all images.
Our framework allows us to adjust this hyperparameter dynamically: if the open vocabulary detector fails, we can lower the threshold and run the visual program again.
In \Cref{fig:selftune}, we can see that variants with a dynamic object detection threshold outperform all other variants and achieve the best performance.
Note that the variant that achieves the highest performance after five trials has a lower performance for the first trial.
This happens because we start with a higher object detection threshold value of 0.15 (by default, we use 0.1).
In this case, initially there will be more false negatives, but also fewer false positives.
As we decrease the threshold in subsequent trials, the previously false negatives are detected and the queries are correctly answered.

\begin{figure}[t]
    \centering
    \begin{subfigure}[t]{0.49\textwidth}
        \centering
        \includegraphics[width=\textwidth]{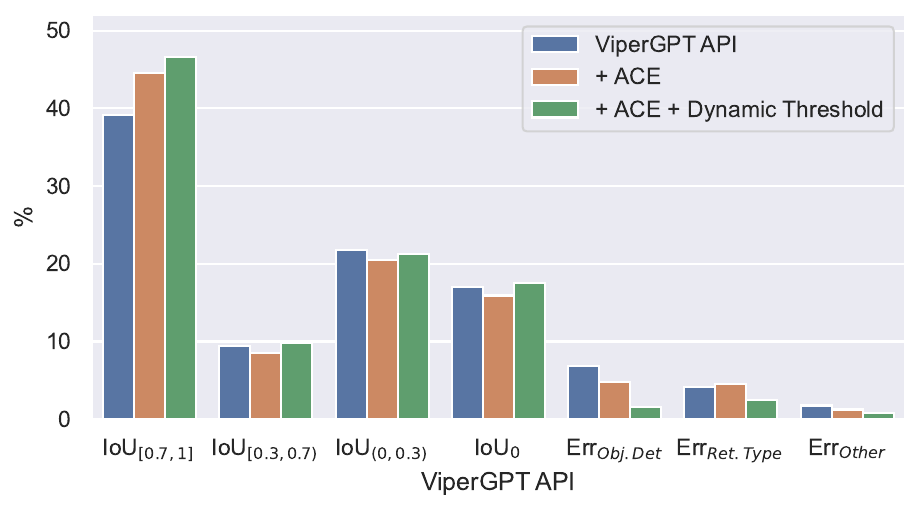}
    \end{subfigure}%
    ~
    \begin{subfigure}[t]{0.49\textwidth}
        \centering
        \includegraphics[width=\textwidth]{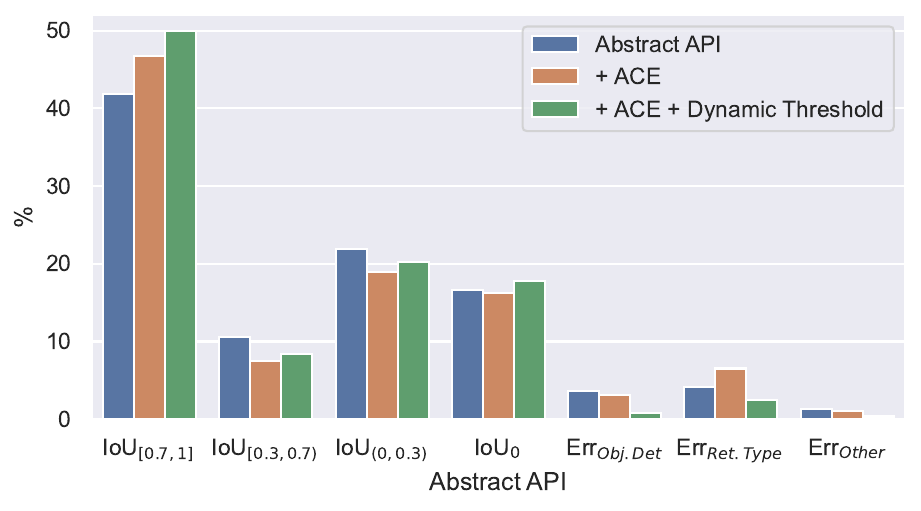}
    \end{subfigure}
    \caption{
    Error diagrams for the ViperGPT API and our Abstract API.
    We visualize the percentages of samples with IoU in certain ranges.
    ``Err'' classes are samples for which code execution failed due to either:
    object detection (Obj.Det), wrong return type (Ret.Type) or some other error (Other) e.g. hallucination.
    }
    \label{fig:error_analysis}
\end{figure}

\subsection{Error analysis}

Another benefit of visual reasoning with LLMs as programmers is interpretability.
For example, we can get insights into the percentage of successful program executions, which can be further decomposed into the ones that resulted in correct or incorrect responses, and for the programs that failed to execute, we can provide further insights into why they failed, i.e. which module failed to return the correct result.
\Cref{fig:error_analysis} shows one such error analysis on RefCOCO.
Categories labeled with ``Error'' are the ones for which code failed to execute due to either object detection (Obj.Det), wrong return type (Ret.Type) or some other error (Other) e.g. hallucination.
For all other cases, code executed correctly (it returned a bounding box), but sometimes it failed to detect the object (``$\text{IoU}=0$'' case).
First, we notice that for both APIs the number of ``correct'' detections (IoU higher than 0.7) grows as we include ACEs and ``self-tuning'' through the dynamic object detection threshold.
We can also see that the percentages of samples with high IoU are always higher for our Abstract API compared to the ViperGPT API.
Finally, note that the percentage of error cases drops from 12.8\% to 4.8\% for the ViperGPT API and from 9.1\% to 3.8\% for our Abstract API.

\section{Related work}

\paragraph{Visual reasoning with end-to-end monolithic models.}
Recently, SotA on VQA has been largely obtained by scaling end-to-end vision and language models (VLMs) in terms of their size, training data, and compute \citep{alayrac2022flamingo,chen2022pali,yu2022coca,wang2022git,gan2022vision,lu2022unified,li2023blip,driess2023palm,chen2023pali3,chen2023paliX}.
One of the earliest VLMs Flamingo \citep{alayrac2022flamingo} used a frozen pretrained language model of up to 70B parameters and a frozen pretrained image encoder with 435M parameters and trained only a ``cross-attention'' module that served as an interface between them.
Since then, efforts have mainly gone into scaling both the image and the language models:
GIT \citep{wang2022git} used a 300M language and scaled the image encoder to 4.8B parameters;
PaLI \citep{chen2022pali} scaled both components jointly, language model to 17B and image encoder to 4B parameters; PaLI-X \citep{chen2023pali3} continued this trend of scaling the total number of parameters to 55B by using an image encoder with 22B parameters; PaLM-E scaled the number of total parameters in the VLM to 562B by integrating the 540B PaLM \citep{chowdhery2022palm} and the 22B Vision Transformer \citep{dosovitskiy2020image,dehghani2023scaling}.
On the other hand, BLIP-2 \citep{li2023blip} achieved SotA performance on various tasks with a 12B VLM and PaLI-3 \cite{chen2023pali3} introduced a significantly smaller VLM with 5B total parameters that achieves competitive performance with SotA models on various VLM benchmarks.
These models are typically pretrained on large amounts of data and then fine-tuned for the best performance on the downstream tasks.
In contrast to this, visual reasoning with LLMs as programmers \citep{gupta2023visual,suris2023vipergpt,subramanian2023modular} does not require any fine-tuning or gradient updates on the downstream tasks.
Moreover, even the largest VLMs struggle on the tasks that require compositional reasoning, the ability to generalize, fine-grained spatial capabilities, and counting \citep{bugliarello2023measuring,hsieh2023sugarcrepe,tschannen2023image}.
Further scaling makes them even more data- and compute-hungry; therefore, it is unclear whether scaling alone can solve these tasks.
Conversely, using LLMs as programmers enables task decomposition into subtasks and holds promise of strong generalization and compositional reasoning.

\paragraph{Visual reasoning with Modular Networks.}
Neural Modular Networks (NMNs) \citep{andreas2016neural,johnson2017inferring,hu2017learning} are an alternative approach to monolithic end-to-end networks and offer a potential route to (compositional) generalization.
They are also typically trained end-to-end, using supervised or reinforcement learning.
NMNs are designed to have several modules, and the hope is that during training, each module will learn a different functionality, which will then be reusable across tasks.
However, these models have a number of drawbacks: the program generation requires hand-tuned parsers, and they require optimization through reinforcement learning (e.g., REINFORCE \citet{williams1992simple}), which is often unstable.
Learning all modules end-to-end hinders their ability to generalize \citep{bahdanau2018systematic} and sometimes leads to a mode ``collapse'', where some modules take over all the work and other modules are never activated, or modules that do not learn intended functionalities \citep{subramanian2020obtaining}.
Furthermore, they sometimes require supervision for program learning, which is difficult to obtain at scale.
On the other hand, visual reasoning with LLMs as programmers mitigates many issues of NMNs: it does not require gradient-based training or finetuning, it is able to incorporate any modules or swap the existing ones, it leverages the strong generalization ability of LLMs, and it generates programs by utilizing the in-context learning ability of LLMs, thereby removing the need for training program generators.
The programs generated by LLM do not have to be domain specific, but can use a common language such as Python that does not require custom interpreters.

\paragraph{Visual reasoning with LLMs as programmers.}
The field of LLMs as controllers for visual reasoning has received a great deal of interest recently.
LLMs as controllers (also known as ``tool use'') approach became prominent in the literature \citep{parisi2022talm,schick2023toolformer}, in particular for structured reasoning in the natural language domain \citep{madaan2022language,wang2023code4struct,gao2023pal,chen2022program}.
In the domain of using LLMs as controllers for visual reasoning, PICa \citep{yang2022empirical} solves a knowledge-based VQA task by first extracting an object and captions from the image and then querying GPT-3 with this information and in-context examples to answer a question.
Socratic models \citep{zeng2022socratic}, HuggingGPT \citep{shen2023hugginggpt}, Societies of Mind \citep{zhuge2023mindstorms}, and many other papers \citep{liu2023llava,zhu2023minigpt,zhang2023llama,li2023videochat,yang2024gpt4tools,xu2022multiinstruct,zhang2023video,gao2023llama,ye2023mplug,yang2023mm,lu2024chameleon,wu2023visual,zhu2023chatgpt} (for a comprehensive survey, see \cite{yin2023survey} or \cite{li2023multimodal}) compose vision and language models to ``communicate'' in a fixed ``protocol'' and solve tasks such as image captioning, visual question answering, image generation, and robot planning.
On the other hand, models such as VisProg \citep{gupta2023visual}, ViperGPT \citep{suris2023vipergpt} and CodeVQA \citep{subramanian2023modular} go beyond a fixed communication ``protocol'' by having LLM write (Python) programs.
During execution, the program calls individual vision modules (such as the object detector and depth estimator) through an \textit{API} that is provided in the prompt.
Additionally, VisProg is also capable of generating images as program output.
These models showed great performance and achieved SotA on tasks such as compositional visual question answering, visual grounding, and video temporal reasoning.
However, in their current form, these models rely on heavy human engineering of query-code examples in the prompt that are dataset- and task-specific and require significant labor by highly skilled workers.
Our framework, on the other hand, is able to automatically generate in-context examples, removing the need for humans to write query-code examples, uses an Abstract API that puts less burden on the LLM to have strong spatial and temporal reasoning abilities, and proposes a way to ``self-correct'' the programs that failed to execute.
Note that our framework is equally applicable to all of the above approaches.

\paragraph{Automatizing prompt engineering.}
Vast literature shows that prompt format and contents are often important for achieving good performance with an LLM \citep{reynolds2021prompt,zhao2021calibrate,lu2021fantastically,moradi2021evaluating,madaan2022text,wei2023larger}.
A prompt typically consists of a task description (in natural language), in-context examples (e.g. query-code in ViperGPT) and an API (in the case where LLMs write a program).
Various prompting techniques have been engineered, such as Chain-of-Thought prompting \citep{wei2022chain}, Self-Consistency \citep{wang2022self}, Tree of Thoughts \citep{yao2023tree}, Graph of Thoughts \citep{besta2023graph}, Plan-and-Solve Prompting \citep{wang2023plan}, Least-to-Most Prompting \citep{zhou2022least}, etc.
All these techniques rely on human prompt engineering, in particular, on in-context examples.
On the other hand, some methods try to automate prompt engineering.
They sometimes use gradient-based optimization \citep{shin2020autoprompt,gao2020making,wen2023hard} and some approaches require only API access to the model \citep{xu2022gps,prasad2022grips}.
Other works use LLMs for prompt optimization.
APE \citep{zhou2022large} first generates instructions with an LLM, then selects instructions with the highest accuracy, and uses them for future LLM prompts.
APO \citep{pryzant2023automatic} generates feedback with an LLM that informs how to update the previous instruction.
OPRO \citep{yang2023large} uses an LLM to generate new instructions at each optimization step, asking the LLM to improve task accuracy by changing task instructions, which requires determining the score on a small set of labeled examples and providing it in the meta-prompt.
Promptbreeder \citep{fernando2023promptbreeder} goes a step further and proposes a self-referential self-improvement LLM using a meta-prompt that controls the generation of the main (task) prompt and evolves both via mutation.
More importantly, Promptbreeder shows some surprising results such that a simple prompt ``SOLUTION'' outperforms all previous approaches.
This further demonstrates the sensitivity of LLMs and the importance of automatizing the prompt engineering process.
Common to all above frameworks for automatizing prompting is that they automatize the ``task description'' part of the prompt.
On the other hand, in our framework, we automatize the generation of in-context examples, which might have an even greater influence on the performance of the LLM.

\paragraph{LLMs and self-correction.}
In the LLM literature, there have been mixed findings on the ability of LLMs to critique and self-correct their own reasoning and outputs.
Self-Refine \citep{madaan2023self} provides feedback to the LLM of the previously generated output, which is then refined.
Several other approaches show benefits of providing feedback to LLM in improving reasoning \citep{shinn2024reflexion, madaan2023self}, code generation \citep{chen2023teaching,olausson2023demystifying,chen2023improving}, improving LLM alignment \citep{bai2022constitutional,ganguli2023capacity}, etc.
On the other hand, there has been increasing evidence that LLMs cannot self-correct reasoning yet \citep{huang2023large,stechly2023gpt4,valmeekam2023large}, unless they receive external feedback, which usually requires access to ground truth labels.
In our work, we also found that providing the previous question and code as feedback to the model did not improve the results.
However, we show that it is possible to improve performance by tuning hyperparameters on the fly, a direction that, to the best of our knowledge, has not been explored previously.

\section{Discussion and future work}

Although the LLMs as controllers framework is very promising for visual reasoning, there is much future work to be explored.
First, the use of video-specific models (or tools) could greatly improve performance on video tasks compared to the image-specific models we used.
Moreover, the code generating LLM currently only takes the question as the input, but for some questions the program that correctly solves the question can only be generated given the image or video as the input too.

The results with the Abstract API show that this is a promising path forward, but more research is needed to find the ``optimal'' set of visual and temporal routines.
Starting from these primitive routines, the model should be able to build an ever-growing library of routines (e.g. through the ACE generation process) that it can later reuse.
This growing library of routines will most likely grow larger than the size of the context window, so research is needed on an API ``router'' that can select routines that are relevant to a specific task at hand.
Furthermore, it would be important to research ways of eliminating the need for few-shot examples when generating ACEs, e.g. by providing a natural language dataset specification (a datasheet).

Lastly, more effort should be put into creating better benchmarks for evaluating compositional visual reasoning, as current ones have a number of limitations.
For example, not all samples in RefCOCO and RefCOCO+ require compositional reasoning, so the LLM should only query the open-vocabulary object detector.
Similarly, many referring expressions in GQA are ambiguous in the sense that there is not a single unique answer.
Finally, NExT-QA contains ambiguous questions (e.g. why someone did certain actions) or questions that can be answered by looking at the multiple choice answers only and disregarding the visual input altogether.
The Perception Test \citep{puatruaucean2023perception} is a promising benchmark for future work, as it was specifically created to avoid such problems.
We hope that our findings inform future research on LLMs as controllers for visual reasoning and encourage systematic evaluation and benchmarking efforts in the future.

\section{Conclusion}

In this work, we present a framework that makes LLMs as programmers for visual reasoning more robust, removes the need for human engineering of in-context examples (ICEs), and thus brings them a step closer to \textit{truly} zero-shot visual reasoners.
We introduce an ``Abstract API'' that consists of spatially and temporally abstract routines, which improves performance by reducing the burden on the code-generating LLM to have strong spatial and temporal reasoning.
By using a few labeled examples, we show how one can generate query-code ICEs automatically (ACEs) in a zero-shot manner.
When used as in-context examples, ACEs consistently improve performance, eliminating the need for human engineering of ICEs.
We demonstrate how LLMs as controllers for visual reasoning can (to a certain extent) perform ``self-correction'' through ``self-debugging'' and ``self-tuning'' without any ground-truth labels.
In self-debugging, generating new code from scratch led to consistently better results, but providing the previous query-code pair as feedback to the LLM did not improve performance.
In self-tuning, we show that the object detector threshold hyperparameter can be tuned automatically if code execution fails due to this module.
Across a number of compositional question-answering and video temporal reasoning tasks, we demonstrate that each component of our framework consistently leads to improvement.

\bibliography{main}
\bibliographystyle{tmlr}

\clearpage

\appendix
\section{Appendix}

\subsection{Ablations}
\label{sec:ablations}

Here we present hyperparameter ablations, namely over the code-generating LLM (\texttt{code-bison}) temperature and the open-vocabulary object detector (OWLv2) threshold.

\begin{table}[h]
\caption{
Code-generating LLM (\texttt{code-bison}) temperature hyperparameter ablations (with ACEs).
The scores are reported as mean $\pm$ standard deviation across three random seeds.
}
\label{tbl:ablation-llm}
\begin{center}
\begin{tabular}{lcccc}
\toprule
Model        & LLM temp. & RefCOCO (IoU)                & RefCOCO+ (IoU)               & NExT-QA (acc.) \\
\midrule
ViperGPT API & 0.0       & 41.4 {\scriptsize $\pm$ 0.3} & 39.8 {\scriptsize $\pm$ 0.0} & 36.0 {\scriptsize $\pm$ 0.1} \\
             & 0.4       & 46.9 {\scriptsize $\pm$ 0.8} & 41.7 {\scriptsize $\pm$ 0.4} & 53.1 {\scriptsize $\pm$ 0.1} \\
             & 0.8       & 46.9 {\scriptsize $\pm$ 0.3} & 42.0 {\scriptsize $\pm$ 0.7} & 53.2 {\scriptsize $\pm$ 0.1} \\
             & 1.0       & 46.2 {\scriptsize $\pm$ 0.7} & 40.9 {\scriptsize $\pm$ 0.6} & 50.3 {\scriptsize $\pm$ 0.5} \\
\midrule
Abstract API & 0.0       & 47.0 {\scriptsize $\pm$ 0.1} & 41.9 {\scriptsize $\pm$ 0.3} & 59.1 {\scriptsize $\pm$ 0.0} \\
             & 0.4       & 48.2 {\scriptsize $\pm$ 0.1} & \textbf{44.7} {\scriptsize $\pm$ 0.2} & \textbf{61.0} {\scriptsize $\pm$ 0.4} \\
             & 0.8       & 48.2 {\scriptsize $\pm$ 0.0} & 43.0 {\scriptsize $\pm$ 0.2} & 60.6 {\scriptsize $\pm$ 0.6} \\
             & 1.0       & \textbf{48.8} {\scriptsize $\pm$ 0.1} & 42.8 {\scriptsize $\pm$ 0.3} & 59.9 {\scriptsize $\pm$ 0.7} \\
\bottomrule
\end{tabular}
\end{center}
\end{table}

In \Cref{tbl:ablation-llm}, we report the scores for different \texttt{code-bison} LLM temperatures: 0, 0.4, 0.8 and 1.0.
We found the deterministic case to underperform compared to the cases with a temperature higher than zero.
This indicates that the solutions of which the models are most ``confident'' of are not necessarily always correct.
On the other hand, when the temperature is too high, the model starts to hallucinate functions that do not exist in the API and the performance degrades.
Early in our work, we settled on the \texttt{code-bison} LLM temperature of 0.4 and did not tune it further.

\begin{table}[h]
\caption{
Open-vocabulary object detector (OWLv2) threshold hyperparameter ablations.
The scores are reported as mean $\pm$ standard deviation across three random seeds.
}
\label{tbl:ablation-owlv2}
\begin{center}
\begin{tabular}{lcccc}
\toprule
Model                          & OWLv2 thrs. & RefCOCO (IoU)                & RefCOCO+ (IoU)               \\
\midrule
\multirow{ 2}{*}{ViperGPT API} & 0.05        & 42.6 {\scriptsize $\pm$ 0.3} & 41.9 {\scriptsize $\pm$ 0.3} \\
                               & 0.10        & 46.9 {\scriptsize $\pm$ 0.8} & 41.7 {\scriptsize $\pm$ 0.4} \\
                               & 0.15        & 45.0 {\scriptsize $\pm$ 0.5} & 38.2 {\scriptsize $\pm$ 0.2} \\
                               & 0.20        & 33.3 {\scriptsize $\pm$ 0.6} & 29.8 {\scriptsize $\pm$ 0.2} \\
\midrule
\multirow{ 2}{*}{Abstract API} & 0.05        & 42.8 {\scriptsize $\pm$ 0.3} & 42.7 {\scriptsize $\pm$ 0.7} \\
                               & 0.10        & \textbf{48.2} {\scriptsize $\pm$ 0.1} & \textbf{44.7} {\scriptsize $\pm$ 0.2} \\
                               & 0.15        & 47.1 {\scriptsize $\pm$ 0.2} & 38.6 {\scriptsize $\pm$ 0.1} \\
                               & 0.20        & 33.7 {\scriptsize $\pm$ 0.4} & 30.1 {\scriptsize $\pm$ 0.4} \\
\bottomrule
\end{tabular}
\end{center}
\end{table}

\Cref{tbl:ablation-owlv2} shows the effect of using different thresholds for OWLv2 open vocabulary detector.
This threshold controls the level of `sensitivity' of the open vocabulary detector.
If the threshold value is set too high, we will have fewer false positives, but also more false negatives.
We perform this study on RefCOCO and RefCOCO+.
On both datasets, the threshold of 0.1 achieves the best results, so by default we use this threshold in all our experiments.

\begin{table}[h]
\caption{
Results on RefCOCO with randomly sampled few-shot samples for generating ACEs.
Shown are IoU scores for zero-shot (w/o ACE) setting, with the default setting that we used throughout the paper (16 manually sampled few-shot samples from the dataset (16 manual)) and with 100, 50 and 16 randomly-sampled samples (100, 50 and 16 random) from the dataset.
The IoU scores are reported as mean $\pm$ standard deviation across three random seeds.
}
\label{tbl:random-few-shot-samples-refcoco}
\begin{center}
\begin{tabular}{lccccc}
\toprule
API      & w/o ACE & 16 manual & 100 random & 50 random & 16 random \\
\midrule
ViperGPT &  41.7 {\scriptsize $\pm$ 0.6} &  46.9 {\scriptsize $\pm$ 0.8} &  47.9 {\scriptsize $\pm$ 0.2} &  45.8 {\scriptsize $\pm$ 0.4} &  41.5 {\scriptsize $\pm$ 0.2} \\
Abstract &  44.4 {\scriptsize $\pm$ 0.9} &  48.2 {\scriptsize $\pm$ 0.1} &  49.1 {\scriptsize $\pm$ 0.2} &  49.0 {\scriptsize $\pm$ 0.2} &  46.9 {\scriptsize $\pm$ 0.2} \\
\bottomrule
\end{tabular}
\end{center}
\end{table}

\begin{table}[h]
\caption{
Results on RefCOCO+ with randomly sampled few-shot samples for generating ACEs.
Shown are IoU scores for zero-shot (w/o ACE) setting, with the default setting that we used throughout the paper (16 manually sampled few-shot samples from the dataset (16 manual)) and with 100, 50 and 16 randomly-sampled samples (100, 50 and 16 random) from the dataset.
The IoU scores are reported as mean $\pm$ standard deviation across three random seeds.
}
\label{tbl:random-few-shot-samples-refcoco-plus}
\begin{center}
\begin{tabular}{lccccc}
\toprule
API      & w/o ACE & 16 manual & 100 random & 50 random & 16 random \\
\midrule
ViperGPT &  36.7 {\scriptsize $\pm$ 0.4} &  41.7 {\scriptsize $\pm$ 0.4} &  41.3 {\scriptsize $\pm$ 0.1} &  40.3 {\scriptsize $\pm$ 0.3} &  36.3 {\scriptsize $\pm$ 0.1} \\
Abstract &  38.2 {\scriptsize $\pm$ 0.0} &  42.9 {\scriptsize $\pm$ 0.2} &  42.8 {\scriptsize $\pm$ 0.6} &  41.8 {\scriptsize $\pm$ 0.0} &  40.2 {\scriptsize $\pm$ 0.4} \\
\bottomrule
\end{tabular}
\end{center}
\end{table}

In \Cref{tbl:random-few-shot-samples-refcoco} and \Cref{tbl:random-few-shot-samples-refcoco-plus} we show results for RefCOCO and RefCOCO+ when using randomly sampled samples for the generation of ACEs.
For comparison purposes, in the tables we also provide scores when not using any ACEs and with the default setting when generating ACEs with 16 manually selected few-shot examples.
From the tables, we can see that randomly selected 100 or 50 samples perform similarly as when using 16 manual samples.
With 16 random samples we observe a small drop in performance, though we still observe an improvement compared to the setting without any ACEs.
In summary, this shows that the manual labor for generating ACEs can be removed altogether if we already have some labeled examples.

\clearpage
\subsection{Pretrained models}
\label{sec:pretrained_models}

Here we specify the pretrained models we used, and compare them with the ones used in ViperGPT:

\begin{itemize}

    \item Open-vocabulary object detector:
    \begin{itemize}
        \item Ours: OWLv2 \citep{minderer2023scaling}.
        \item ViperGPT: GLIP \cite{li2022grounded} from the official GitHub repository\footnote{https://github.com/microsoft/GLIP}.
    \end{itemize}

    \item Depth estimation model:
    \begin{itemize}
        \item Ours: MiDaS \citep{ranftl2020towards} v2 ``DPT\_Small'' from PyTorch hub\footnote{https://pytorch.org/hub/intelisl\_midas\_v2/}.
        \item ViperGPT: MiDaS \citep{ranftl2020towards} v2 ``DPT\_Large'' version from the PyTorch hub\footnote{https://pytorch.org/hub/intelisl\_midas\_v2/}.
    \end{itemize}

    \item Vision-language captioning model:
    \begin{itemize}
        \item Ours: PaLI-3 \citep{chen2023pali3}.
        \item ViperGPT: BLIP-2 \citep{li2023blip} from the official repository\footnote{https://github.com/salesforce/LAVIS/tree/main/projects/blip2}.
    \end{itemize}

    \item CLIP-style image-text embedding model:
    \begin{itemize}
        \item Ours: SigLiT \citep{zhai2023sigmoid}.
        \item ViperGPT: X-VLM \citep{zeng2021multi} version finetuned for retrieval on MSCOCO from the official repository\footnote{https://github.com/zengyan-97/X-VLM}.
    \end{itemize}

    \item Code-generating LLM:
    \begin{itemize}
        \item Ours: \texttt{code-bison} accessible via the Google Cloud Vertex AI API \citep{google2023codebison}.
        \item ViperGPT: Codex (\texttt{code-davinci-002}) via the official OpenAI Python API\footnote{https://openai.com/blog/openai-api}.
    \end{itemize}

    \item Answer selector (based on context information) LLM for multiple choice questions in NExT-QA:
    \begin{itemize}
        \item Ours: \texttt{code-bison} accessible via the Google Cloud Vertex AI API \citep{google2023codebison}.
        \item ViperGPT: GPT-3 via the official OpenAI Python API\footnote{https://openai.com/blog/openai-api}.
    \end{itemize}
\end{itemize}

\subsection{Self-debugging prompt}
\label{sec:self-debugging-prompt}
\begin{lstlisting}[language=Python]
prompt += f"""

Previously, for the query:
# {query}
you've generated the following code:
{code}

The execution of the above code failed and returned the following error message:
{error}.

Given this, correct the above function, such that it executes correctly and solves the same query.

"""
\end{lstlisting}

\clearpage
\subsection{Prompt listings}
\label{sec:prompt_listings}

\subsubsection{RefCOCO and GQA - ViperGPT API}
\begin{lstlisting}[language=Python]
import math

class ImagePatch:
    """A Python class containing a crop of an image centered around a particular object, as well as relevant information.
    Attributes
    ----------
    cropped_image : array_like
        An array-like of the cropped image taken from the original image.
    left, lower, right, upper : int
        An int describing the position of the (left/lower/right/upper) border of the crop's bounding box in the original image.

    Methods
    -------
    find(object_name: str)->List[ImagePatch]
        Returns a list of new ImagePatch objects containing crops of the image centered around any objects found in the
        image matching the object_name.
    exists(object_name: str)->bool
        Returns True if the object specified by object_name is found in the image, and False otherwise.
    verify_property(property: str)->bool
        Returns True if the property is met, and False otherwise.
    compute_depth()->float
        Returns the median depth of the image crop.
    crop(left: int, lower: int, right: int, upper: int)->ImagePatch
        Returns a new ImagePatch object containing a crop of the image at the given coordinates.
    """

    def __init__(self, image, left: int = None, lower: int = None, right: int = None, upper: int = None):
        """Initializes an ImagePatch object by cropping the image at the given coordinates and stores the coordinates as
        attributes. If no coordinates are provided, the image is left unmodified, and the coordinates are set to the
        dimensions of the image.
        Parameters
        -------
        image : array_like
            An array-like of the original image.
        left, lower, right, upper : int
            An int describing the position of the (left/lower/right/upper) border of the crop's bounding box in the original image.
        """
        if left is None and right is None and upper is None and lower is None:
            self.cropped_image = image
            self.left = 0
            self.lower = 0
            self.right = image.shape[2]  # width
            self.upper = image.shape[1]  # height
        else:
            self.cropped_image = image[:, lower:upper, left:right]
            self.left = left
            self.upper = upper
            self.right = right
            self.lower = lower

        self.width = self.cropped_image.shape[2]
        self.height = self.cropped_image.shape[1]

        self.horizontal_center = (self.left + self.right) / 2
        self.vertical_center = (self.lower + self.upper) / 2

    def find(self, object_name: str) -> List[ImagePatch]:
        """Returns a list of ImagePatch objects matching object_name contained in the crop if any are found.
        Otherwise, returns an empty list.
        Parameters
        ----------
        object_name : str
            the name of the object to be found

        Returns
        -------
        List[ImagePatch]
            a list of ImagePatch objects matching object_name contained in the crop

        Examples
        --------
        >>> # return the foo
        >>> def execute_command(image) -> List[ImagePatch]:
        >>>     image_patch = ImagePatch(image)
        >>>     foo_patches = image_patch.find("foo")
        >>>     return foo_patches
        """
        return find_in_image(self.cropped_image, object_name)

    def exists(self, object_name: str) -> bool:
        """Returns True if the object specified by object_name is found in the image, and False otherwise.
        Parameters
        -------
        object_name : str
            A string describing the name of the object to be found in the image.

        Examples
        -------
        >>> # Are there both foos and garply bars in the photo?
        >>> def execute_command(image)->str:
        >>>     image_patch = ImagePatch(image)
        >>>     is_foo = image_patch.exists("foo")
        >>>     is_garply_bar = image_patch.exists("garply bar")
        >>>     return is_foo and is_garply_bar
        """
        return len(self.find(object_name)) > 0

    def verify_property(self, object_name: str, visual_property: str) -> bool:
        """Returns True if the object possesses the visual property, and False otherwise.
        Differs from 'exists' in that it presupposes the existence of the object specified by object_name, instead checking whether the object possesses the property.
        Parameters
        -------
        object_name : str
            A string describing the name of the object to be found in the image.
        visual_property : str
            A string describing the simple visual property (e.g., color, shape, material) to be checked.

        Examples
        -------
        >>> # Do the letters have blue color?
        >>> def execute_command(image) -> str:
        >>>     image_patch = ImagePatch(image)
        >>>     letters_patches = image_patch.find("letters")
        >>>     # Question assumes only one letter patch
        >>>     return letters_patches[0].verify_property("letters", "blue")
        """
        return verify_property(self.cropped_image, object_name, property)

    def compute_depth(self):
        """Returns the median depth of the image crop
        Parameters
        ----------
        Returns
        -------
        float
            the median depth of the image crop

        Examples
        --------
        >>> # the bar furthest away
        >>> def execute_command(image)->ImagePatch:
        >>>     image_patch = ImagePatch(image)
        >>>     bar_patches = image_patch.find("bar")
        >>>     bar_patches.sort(key=lambda bar: bar.compute_depth())
        >>>     return bar_patches[-1]
        """
        depth_map = compute_depth(self.cropped_image)
        return depth_map.median()

    def crop(self, left: int, lower: int, right: int, upper: int) -> ImagePatch:
        """Returns a new ImagePatch cropped from the current ImagePatch.
        Parameters
        -------
        left, lower, right, upper : int
            The (left/lower/right/upper)most pixel of the cropped image.
        -------
        """
        return ImagePatch(self.cropped_image, left, lower, right, upper)

    def overlaps_with(self, left, lower, right, upper):
        """Returns True if a crop with the given coordinates overlaps with this one,
        else False.
        Parameters
        ----------
        left, lower, right, upper : int
            the (left/lower/right/upper) border of the crop to be checked

        Returns
        -------
        bool
            True if a crop with the given coordinates overlaps with this one, else False

        Examples
        --------
        >>> # black foo on top of the qux
        >>> def execute_command(image) -> ImagePatch:
        >>>     image_patch = ImagePatch(image)
        >>>     qux_patches = image_patch.find("qux")
        >>>     qux_patch = qux_patches[0]
        >>>     foo_patches = image_patch.find("black foo")
        >>>     for foo in foo_patches:
        >>>         if foo.vertical_center > qux_patch.vertical_center
        >>>             return foo
        """
        return self.left <= right and self.right >= left and self.lower <= upper and self.upper >= lower


def best_image_match(list_patches: List[ImagePatch], content: List[str], return_index=False) -> Union[ImagePatch, int]:
    """Returns the patch most likely to contain the content.
    Parameters
    ----------
    list_patches : List[ImagePatch]
    content : List[str]
        the object of interest
    return_index : bool
        if True, returns the index of the patch most likely to contain the object

    Returns
    -------
    int
        Patch most likely to contain the object
    """
    return best_image_match(list_patches, content, return_index)


def distance(patch_a: ImagePatch, patch_b: ImagePatch) -> float:
    """
    Returns the distance between the edges of two ImagePatches. If the patches overlap, it returns a negative distance
    corresponding to the negative intersection over union.

    Parameters
    ----------
    patch_a : ImagePatch
    patch_b : ImagePatch

    Examples
    --------
    # Return the qux that is closest to the foo
    >>> def execute_command(image):
    >>>     image_patch = ImagePatch(image)
    >>>     qux_patches = image_patch.find('qux')
    >>>     foo_patches = image_patch.find('foo')
    >>>     foo_patch = foo_patches[0]
    >>>     qux_patches.sort(key=lambda x: distance(x, foo_patch))
    >>>     return qux_patches[0]
    """
    return distance(patch_a, patch_b)

INSERT_IN_CONTEXT_EXAMPLES_HERE

Write a function using Python and the ImagePatch class (above) that could be executed to provide an answer to the query. 

Consider the following guidelines:
- Use base Python (comparison, sorting) for basic logical operations, left/right/up/down, math, etc.
- Make sure to always return an ImagePatch object.
- Make sure that for all possible control flows, the program always returns an ImagePatch object.

INSERT_PREVIOUS_CODE_AND_ERROR_HERE

# INSERT_QUERY_HERE
\end{lstlisting}

\clearpage
\subsubsection{RefCOCO and GQA - Abstract API}
\begin{lstlisting}[language=Python]
import math

class ImagePatch:
    """A Python class containing a crop of an image centered around a particular object, as well as relevant information.
    Attributes
    ----------
    cropped_image : array_like
        An array-like of the cropped image taken from the original image.
    left, lower, right, upper : int
        An int describing the position of the (left/lower/right/upper) border of the crop's bounding box in the original image.

    Methods
    -------
    find(object_name: str)->List[ImagePatch]
        Returns a list of new ImagePatch objects containing crops of the image centered around any objects found in the
        image matching the object_name.
    exists(object_name: str)->bool
        Returns True if the object specified by object_name is found in the image, and False otherwise.
    verify_property(property: str)->bool
        Returns True if the property is met, and False otherwise.
    compute_depth()->float
        Returns the median depth of the image crop.
    crop(left: int, lower: int, right: int, upper: int)->ImagePatch
        Returns a new ImagePatch object containing a crop of the image at the given coordinates.
    """

    def __init__(self, image, left: int = None, lower: int = None, right: int = None, upper: int = None):
        """Initializes an ImagePatch object by cropping the image at the given coordinates and stores the coordinates as
        attributes. If no coordinates are provided, the image is left unmodified, and the coordinates are set to the
        dimensions of the image.
        Parameters
        -------
        image : array_like
            An array-like of the original image.
        left, lower, right, upper : int
            An int describing the position of the (left/lower/right/upper) border of the crop's bounding box in the original image.
        """
        if left is None and right is None and upper is None and lower is None:
            self.cropped_image = image
            self.left = 0
            self.lower = 0
            self.right = image.shape[2]  # width
            self.upper = image.shape[1]  # height
        else:
            self.cropped_image = image[:, lower:upper, left:right]
            self.left = left
            self.upper = upper
            self.right = right
            self.lower = lower

        self.width = self.cropped_image.shape[2]
        self.height = self.cropped_image.shape[1]

        self.horizontal_center = (self.left + self.right) / 2
        self.vertical_center = (self.lower + self.upper) / 2

    def find(self, object_name: str) -> List[ImagePatch]:
        """Returns a list of ImagePatch objects matching object_name contained in the crop if any are found.
        Otherwise, returns an empty list.
        Parameters
        ----------
        object_name : str
            the name of the object to be found

        Returns
        -------
        List[ImagePatch]
            a list of ImagePatch objects matching object_name contained in the crop

        Examples
        --------
        >>> # return the foo
        >>> def execute_command(image) -> List[ImagePatch]:
        >>>     image_patch = ImagePatch(image)
        >>>     foo_patches = image_patch.find("foo")
        >>>     return foo_patches
        """
        return find_in_image(self.cropped_image, object_name)

    def exists(self, object_name: str) -> bool:
        """Returns True if the object specified by object_name is found in the image, and False otherwise.
        Parameters
        -------
        object_name : str
            A string describing the name of the object to be found in the image.

        Examples
        -------
        >>> # Are there both foos and garply bars in the photo?
        >>> def execute_command(image)->str:
        >>>     image_patch = ImagePatch(image)
        >>>     is_foo = image_patch.exists("foo")
        >>>     is_garply_bar = image_patch.exists("garply bar")
        >>>     return is_foo and is_garply_bar
        """
        return len(self.find(object_name)) > 0

    def verify_property(self, object_name: str, visual_property: str) -> bool:
        """Returns True if the object possesses the visual property, and False otherwise.
        Differs from 'exists' in that it presupposes the existence of the object specified by object_name, instead checking whether the object possesses the property.
        Parameters
        -------
        object_name : str
            A string describing the name of the object to be found in the image.
        visual_property : str
            A string describing the simple visual property (e.g., color, shape, material) to be checked.

        Examples
        -------
        >>> # Do the letters have blue color?
        >>> def execute_command(image) -> str:
        >>>     image_patch = ImagePatch(image)
        >>>     letters_patches = image_patch.find("letters")
        >>>     # Question assumes only one letter patch
        >>>     return letters_patches[0].verify_property("letters", "blue")
        """
        return verify_property(self.cropped_image, object_name, property)

    def compute_depth(self):
        """Returns the median depth of the image crop
        Parameters
        ----------
        Returns
        -------
        float
            the median depth of the image crop

        Examples
        --------
        >>> # the bar furthest away
        >>> def execute_command(image)->ImagePatch:
        >>>     image_patch = ImagePatch(image)
        >>>     bar_patches = image_patch.find("bar")
        >>>     bar_patches.sort(key=lambda bar: bar.compute_depth())
        >>>     return bar_patches[-1]
        """
        depth_map = compute_depth(self.cropped_image)
        return depth_map.median()

    def crop(self, left: int, lower: int, right: int, upper: int) -> ImagePatch:
        """Returns a new ImagePatch cropped from the current ImagePatch.
        Parameters
        -------
        left, lower, right, upper : int
            The (left/lower/right/upper)most pixel of the cropped image.
        -------
        """
        return ImagePatch(self.cropped_image, left, lower, right, upper)

    def overlaps_with(self, left, lower, right, upper):
        """Returns True if a crop with the given coordinates overlaps with this one,
        else False.
        Parameters
        ----------
        left, lower, right, upper : int
            the (left/lower/right/upper) border of the crop to be checked

        Returns
        -------
        bool
            True if a crop with the given coordinates overlaps with this one, else False

        Examples
        --------
        >>> # black foo on top of the qux
        >>> def execute_command(image) -> ImagePatch:
        >>>     image_patch = ImagePatch(image)
        >>>     qux_patches = image_patch.find("qux")
        >>>     qux_patch = qux_patches[0]
        >>>     foo_patches = image_patch.find("black foo")
        >>>     for foo in foo_patches:
        >>>         if foo.vertical_center > qux_patch.vertical_center
        >>>             return foo
        """
        return self.left <= right and self.right >= left and self.lower <= upper and self.upper >= lower


def best_image_match(list_patches: List[ImagePatch], content: List[str], return_index=False) -> Union[ImagePatch, int]:
    """Returns the patch most likely to contain the content.
    Parameters
    ----------
    list_patches : List[ImagePatch]
    content : List[str]
        the object of interest
    return_index : bool
        if True, returns the index of the patch most likely to contain the object

    Returns
    -------
    int
        Patch most likely to contain the object
    """
    return best_image_match(list_patches, content, return_index)


def distance(patch_a: ImagePatch, patch_b: ImagePatch) -> float:
    """
    Returns the distance between the edges of two ImagePatches. If the patches overlap, it returns a negative distance
    corresponding to the negative intersection over union.

    Parameters
    ----------
    patch_a : ImagePatch
    patch_b : ImagePatch

    Examples
    --------
    # Return the qux that is closest to the foo
    >>> def execute_command(image):
    >>>     image_patch = ImagePatch(image)
    >>>     qux_patches = image_patch.find('qux')
    >>>     foo_patches = image_patch.find('foo')
    >>>     foo_patch = foo_patches[0]
    >>>     qux_patches.sort(key=lambda x: distance(x, foo_patch))
    >>>     return qux_patches[0]
    """
    return distance(patch_a, patch_b)

def get_patch_left_of(patch: ImagePatch) -> ImagePatch:
    left_patch = get_patch_left_of(patch)
    return left_patch

def get_patch_right_of(patch: ImagePatch) -> ImagePatch:
    right_patch = get_patch_right_of(patch)
    return right_patch

def get_patch_above_of(patch: ImagePatch) -> ImagePatch:
    above_patch = get_patch_above_of(patch)
    return above_patch

def get_patch_below_of(patch: ImagePatch) -> ImagePatch:
    below_patch = get_patch_below_of(patch)
    return below_patch

def get_patch_around_of(patch: ImagePatch) -> ImagePatch:
    around_patch = get_patch_around_of(patch)
    return around_patch

def sort_patches_left_to_right(list_patches: List[ImagePatch]) -> List[ImagePatch]:
    """
    Sorts patches according to their horizontal centers.

    Parameters
    ----------
    list_patches : List[ImagePatch]

    Examples
    --------
    # Right foo
    >>> def execute_command(image):
    >>>     image_patch = ImagePatch(image)
    >>>     foo_patches = image_patch.find('foo')
    >>>     foo_patches = sort_patches_left_to_right(foo_patches)
    >>>     right_foo_patch = foo_patches[-1]
    >>>     return right_foo_patch
    """
    return sort_patches_left_to_right(list_patches)


def sort_patches_bottom_to_top(list_patches: List[ImagePatch]) -> List[ImagePatch]:
    """
    Sorts patches according to their vertical centers.

    Parameters
    ----------
    list_patches : List[ImagePatch]

    Examples
    --------
    # Second bar from the top
    >>> def execute_command(image):
    >>>     image_patch = ImagePatch(image)
    >>>     bar_patches = image_patch.find('bar')
    >>>     bar_patches = sort_patches_bottom_to_top(bar_patches)
    >>>     second_topmost_bar_patch = bar_patches[-2]
    >>>     return second_topmost_bar_patch
    """
    return sort_patches_bottom_to_top(list_patches)


def sort_patches_front_to_back(list_patches: List[ImagePatch]) -> List[ImagePatch]:
    """
    Sorts patches according to how far from camera they are.

    Parameters
    ----------
    list_patches : List[ImagePatch]

    Examples
    --------
    # Person in the back
    >>> def execute_command(image):
    >>>     image_patch = ImagePatch(image)
    >>>     person_patches = image_patch.find('person')
    >>>     person_patches = sort_patches_front_to_back(person_patches)
    >>>     person_in_the_back = person_patches[-1]
    >>>     return person_in_the_back
    """
    return sort_patches_front_to_back(list_patches)


def get_middle_patch(list_patches: List[ImagePatch]) -> ImagePatch:
    """
    Returns the middle patch.

    Parameters
    ----------
    list_patches : List[ImagePatch]

    Examples
    --------
    # Middle ham
    >>> def execute_command(image):
    >>>     image_patch = ImagePatch(image)
    >>>     ham_patches = image_patch.find('ham')
    >>>     middle_ham_patch = get_middle_patch(ham_patches)
    >>>     return middle_ham_patch
    """
    return get_middle_patch(list_patches)


def get_patch_closest_to_anchor_object(list_patches: List[ImagePatch], anchor_object: ImagePatch) -> ImagePatch:
    """
    Returns the object from list_patches that is the closest to the anchor_object.

    Parameters
    ----------
    list_patches : List[ImagePatch]
    anchor_object : ImagePatch

    Examples
    --------
    # Foo next to bar
    >>> def execute_command(image):
    >>>     image_patch = ImagePatch(image)
    >>>     foo_patches = image_patch.find('foo')
    >>>     bar_patches = image_patch.find('bar')
    >>>     bar_patch = bar_patches[0]
    >>>     foo_next_to_bar_patch = get_patch_closest_to_anchor_object(foo_patches, bar_patch)
    >>>     return foo_next_to_bar_patch
    """
    return get_patch_closest_to_anchor_object(list_patches, anchor_object)


INSERT_IN_CONTEXT_EXAMPLES_HERE

Write a function using Python and the ImagePatch class (above) that could be executed to provide an answer to the query. 

Consider the following guidelines:
- Use base Python (comparison, sorting) for basic logical operations, left/right/up/down, math, etc.
- Make sure to always return an ImagePatch object.
- Make sure that for all possible control flows, the program always returns an ImagePatch object.
- ImagePatch class uses left and right to denote horizontal edges.
- ImagePatch class uses bottom and top to denote vertical edges.

INSERT_PREVIOUS_CODE_AND_ERROR_HERE

# INSERT_QUERY_HERE
\end{lstlisting}

\clearpage
\subsubsection{NExT-QA - ViperGPT API}
\begin{lstlisting}[language=Python]
import math

class ImagePatch:
    """A Python class containing a crop of an image centered around a particular object, as well as relevant information.
    Attributes
    ----------
    cropped_image : array_like
        An array-like of the cropped image taken from the original image.
    left, lower, right, upper : int
        An int describing the position of the (left/lower/right/upper) border of the crop's bounding box in the original image.

    Methods
    -------
    find(object_name: str)->List[ImagePatch]
        Returns a list of new ImagePatch objects containing crops of the image centered around any objects found in the
        image matching the object_name.
    exists(object_name: str)->bool
        Returns True if the object specified by object_name is found in the image, and False otherwise.
    verify_property(property: str)->bool
        Returns True if the property is met, and False otherwise.
    best_text_match(option_list: List[str], prefix: str)->str
        Returns the string that best matches the image.
    simple_query(question: str=None)->str
        Returns the answer to a basic question asked about the image. If no question is provided, returns the answer to "What is this?".
    compute_depth()->float
        Returns the median depth of the image crop.
    crop(left: int, lower: int, right: int, upper: int)->ImagePatch
        Returns a new ImagePatch object containing a crop of the image at the given coordinates.
    """

    def __init__(self, image, left: int = None, lower: int = None, right: int = None, upper: int = None):
        """Initializes an ImagePatch object by cropping the image at the given coordinates and stores the coordinates as
        attributes. If no coordinates are provided, the image is left unmodified, and the coordinates are set to the
        dimensions of the image.
        Parameters
        -------
        image : array_like
            An array-like of the original image.
        left, lower, right, upper : int
            An int describing the position of the (left/lower/right/upper) border of the crop's bounding box in the original image.
        """
        if left is None and right is None and upper is None and lower is None:
            self.cropped_image = image
            self.left = 0
            self.lower = 0
            self.right = image.shape[2]  # width
            self.upper = image.shape[1]  # height
        else:
            self.cropped_image = image[:, lower:upper, left:right]
            self.left = left
            self.upper = upper
            self.right = right
            self.lower = lower

        self.width = self.cropped_image.shape[2]
        self.height = self.cropped_image.shape[1]

        self.horizontal_center = (self.left + self.right) / 2
        self.vertical_center = (self.lower + self.upper) / 2

    def find(self, object_name: str) -> List[ImagePatch]:
        """Returns a list of ImagePatch objects matching object_name contained in the crop if any are found.
        Otherwise, returns an empty list.
        Parameters
        ----------
        object_name : str
            the name of the object to be found

        Returns
        -------
        List[ImagePatch]
            a list of ImagePatch objects matching object_name contained in the crop

        Examples
        --------
        >>> # return the foo
        >>> def execute_command(image) -> List[ImagePatch]:
        >>>     image_patch = ImagePatch(image)
        >>>     foo_patches = image_patch.find("foo")
        >>>     return foo_patches
        """
        return find_in_image(self.cropped_image, object_name)

    def exists(self, object_name: str) -> bool:
        """Returns True if the object specified by object_name is found in the image, and False otherwise.
        Parameters
        -------
        object_name : str
            A string describing the name of the object to be found in the image.

        Examples
        -------
        >>> # Are there both foos and garply bars in the photo?
        >>> def execute_command(image)->str:
        >>>     image_patch = ImagePatch(image)
        >>>     is_foo = image_patch.exists("foo")
        >>>     is_garply_bar = image_patch.exists("garply bar")
        >>>     return bool_to_yesno(is_foo and is_garply_bar)
        """
        return len(self.find(object_name)) > 0

    def verify_property(self, object_name: str, visual_property: str) -> bool:
        """Returns True if the object possesses the visual property, and False otherwise.
        Differs from 'exists' in that it presupposes the existence of the object specified by object_name, instead checking whether the object possesses the property.
        Parameters
        -------
        object_name : str
            A string describing the name of the object to be found in the image.
        visual_property : str
            A string describing the simple visual property (e.g., color, shape, material) to be checked.

        Examples
        -------
        >>> # Do the letters have blue color?
        >>> def execute_command(image) -> str:
        >>>     image_patch = ImagePatch(image)
        >>>     letters_patches = image_patch.find("letters")
        >>>     # Question assumes only one letter patch
        >>>     return bool_to_yesno(letters_patches[0].verify_property("letters", "blue"))
        """
        return verify_property(self.cropped_image, object_name, property)

    def best_text_match(self, option_list: List[str]) -> str:
        """Returns the string that best matches the image.
        Parameters
        -------
        option_list : str
            A list with the names of the different options
        prefix : str
            A string with the prefixes to append to the options

        Examples
        -------
        >>> # Is the foo gold or white?
        >>> def execute_command(image)->str:
        >>>     image_patch = ImagePatch(image)
        >>>     foo_patches = image_patch.find("foo")
        >>>     # Question assumes one foo patch
        >>>     return foo_patches[0].best_text_match(["gold", "white"])
        """
        return best_text_match(self.cropped_image, option_list)

    def simple_query(self, question: str = None) -> str:
        """Returns the answer to a basic question asked about the image. If no question is provided, returns the answer
        to "What is this?". The questions are about basic perception, and are not meant to be used for complex reasoning
        or external knowledge.
        Parameters
        -------
        question : str
            A string describing the question to be asked.

        Examples
        -------

        >>> # Which kind of baz is not fredding?
        >>> def execute_command(image) -> str:
        >>>     image_patch = ImagePatch(image)
        >>>     baz_patches = image_patch.find("baz")
        >>>     for baz_patch in baz_patches:
        >>>         if not baz_patch.verify_property("baz", "fredding"):
        >>>             return baz_patch.simple_query("What is this baz?")

        >>> # What color is the foo?
        >>> def execute_command(image) -> str:
        >>>     image_patch = ImagePatch(image)
        >>>     foo_patches = image_patch.find("foo")
        >>>     foo_patch = foo_patches[0]
        >>>     return foo_patch.simple_query("What is the color?")

        >>> # Is the second bar from the left quuxy?
        >>> def execute_command(image) -> str:
        >>>     image_patch = ImagePatch(image)
        >>>     bar_patches = image_patch.find("bar")
        >>>     bar_patches.sort(key=lambda x: x.horizontal_center)
        >>>     bar_patch = bar_patches[1]
        >>>     return bar_patch.simple_query("Is the bar quuxy?")
        """
        return simple_query(self.cropped_image, question)

    def compute_depth(self):
        """Returns the median depth of the image crop
        Parameters
        ----------
        Returns
        -------
        float
            the median depth of the image crop

        Examples
        --------
        >>> # the bar furthest away
        >>> def execute_command(image)->ImagePatch:
        >>>     image_patch = ImagePatch(image)
        >>>     bar_patches = image_patch.find("bar")
        >>>     bar_patches.sort(key=lambda bar: bar.compute_depth())
        >>>     return bar_patches[-1]
        """
        depth_map = compute_depth(self.cropped_image)
        return depth_map.median()

    def crop(self, left: int, lower: int, right: int, upper: int) -> ImagePatch:
        """Returns a new ImagePatch cropped from the current ImagePatch.
        Parameters
        -------
        left, lower, right, upper : int
            The (left/lower/right/upper)most pixel of the cropped image.
        -------
        """
        return ImagePatch(self.cropped_image, left, lower, right, upper)

    def overlaps_with(self, left, lower, right, upper):
        """Returns True if a crop with the given coordinates overlaps with this one,
        else False.
        Parameters
        ----------
        left, lower, right, upper : int
            the (left/lower/right/upper) border of the crop to be checked

        Returns
        -------
        bool
            True if a crop with the given coordinates overlaps with this one, else False

        Examples
        --------
        >>> # black foo on top of the qux
        >>> def execute_command(image) -> ImagePatch:
        >>>     image_patch = ImagePatch(image)
        >>>     qux_patches = image_patch.find("qux")
        >>>     qux_patch = qux_patches[0]
        >>>     foo_patches = image_patch.find("black foo")
        >>>     for foo in foo_patches:
        >>>         if foo.vertical_center > qux_patch.vertical_center
        >>>             return foo
        """
        return self.left <= right and self.right >= left and self.lower <= upper and self.upper >= lower


class VideoSegment:
    """A Python class containing a set of frames represented as ImagePatch objects, as well as relevant information.
    Attributes
    ----------
    video : torch.Tensor
    A tensor of the original video.
    start : int
    An int describing the starting frame in this video segment with respect to the original video.
    end : int
    An int describing the ending frame in this video segment with respect to the original video.
    num_frames->int
    An int containing the number of frames in the video segment.

    Methods
    -------
    frame_iterator->Iterator[ImagePatch]
    trim(start, end)->VideoSegment
    Returns a new VideoSegment containing a trimmed version of the original video at the [start, end] segment.
    select_answer(info, question, options)->str
    Returns the answer to the question given the options and additional information.
    """

    def __init__(self, video: torch.Tensor, start: int = None, end: int = None, parent_start=0, queues=None):
        """Initializes a VideoSegment object by trimming the video at the given [start, end] times and stores the
        start and end times as attributes. If no times are provided, the video is left unmodified, and the times are
        set to the beginning and end of the video.

        Parameters
        -------
        video : torch.Tensor
        A tensor of the original video.
        start : int
        An int describing the starting frame in this video segment with respect to the original video.
        end : int
        An int describing the ending frame in this video segment with respect to the original video.
        """

        if start is None and end is None:
            self.trimmed_video = video
            self.start = 0
            self.end = video.shape[0] # duration
        else:
            self.trimmed_video = video[start:end]
        if start is None:
            start = 0
        if end is None:
            end = video.shape[0]
        self.start = start + parent_start
        self.end = end + parent_start

        self.num_frames = self.trimmed_video.shape[0]

    def frame_iterator(self) -> Iterator[ImagePatch]:
        """Returns an iterator over the frames in the video segment."""
        for i in range(self.num_frames):
            yield ImagePatch(self.trimmed_video[i], self.start + i)

    def trim(self, start: Union[int, None] = None, end: Union[int, None] = None) -> VideoSegment:
        """Returns a new VideoSegment containing a trimmed version of the original video at the [start, end]
        segment.

        Parameters
        ----------
        start : Union[int, None]
        An int describing the starting frame in this video segment with respect to the original video.
        end : Union[int, None]
        An int describing the ending frame in this video segment with respect to the original video.

        Examples
        --------
        >>> # Return the second half of the video
        >>> def execute_command(video):
        >>> video_segment = VideoSegment(video)
        >>> video_second_half = video_segment.trim(video_segment.num_frames // 2, video_segment.num_frames)
        >>> return video_second_half
        """
        if start is not None:
            start = max(start, 0)
        if end is not None:
            end = min(end, self.num_frames)

        return VideoSegment(self.trimmed_video, start, end, self.start)

    def select_answer(self, info: dict, question: str, options: List[str]) -> str:
        return select_answer(self.trimmed_video, info, question, options)

    def __repr__(self):
        return "VideoSegment({}, {})".format(self.start, self.end)

    def simple_query(self, question) -> str:
        """Ask a simple question about the video.

        Examples
        --------
        # why does X happen?
        # possible_answers: ['answer1', 'answer2', 'answer3', 'answer4', 'answer5']
        def execute_command(video, question, possible_answers)->[str, dict]:
            # Create a video segment object
            video_segment = VideoSegment(video)
            # The question is simple, so just ask
            info = video_segment.simple_query("why does X happen?")
            # Choose the answer among given possible answers
            answer = select_answer(info, question, possible_answers)
            return answer
        """
        answer = simple_query(question)
        return answer


def best_image_match(list_patches: List[ImagePatch], content: List[str], return_index=False) -> Union[ImagePatch, int]:
    """Returns the patch most likely to contain the content.
    Parameters
    ----------
    list_patches : List[ImagePatch]
    content : List[str]
        the object of interest
    return_index : bool
        if True, returns the index of the patch most likely to contain the object

    Returns
    -------
    int
        Patch most likely to contain the object
    """
    return best_image_match(list_patches, content, return_index)


def distance(patch_a: ImagePatch, patch_b: ImagePatch) -> float:
    """
    Returns the distance between the edges of two ImagePatches. If the patches overlap, it returns a negative distance
    corresponding to the negative intersection over union.

    Parameters
    ----------
    patch_a : ImagePatch
    patch_b : ImagePatch

    Examples
    --------
    # Return the qux that is closest to the foo
    >>> def execute_command(image):
    >>>     image_patch = ImagePatch(image)
    >>>     qux_patches = image_patch.find('qux')
    >>>     foo_patches = image_patch.find('foo')
    >>>     foo_patch = foo_patches[0]
    >>>     qux_patches.sort(key=lambda x: distance(x, foo_patch))
    >>>     return qux_patches[0]
    """
    return distance(patch_a, patch_b)


def bool_to_yesno(bool_answer: bool) -> str:
    return "yes" if bool_answer else "no"


def select_answer(info: str, question: question, possible_answers: str) -> str:
    """Given an info, question and possible answers, select the correct answer.

    Examples
    --------
    # what does man do at the end of the video
    # possible_answers: ['answer1', 'answer2', 'answer3', 'answer4', 'answer5']
    def execute_command(video, question, possible_answers)->[str, dict]:
        # Create a video segment object
        video_segment = VideoSegment(video)
        # Caption last frame of the video (end of video)
        last_frame = ImagePatch(video_segment, -1)
        last_caption = last_frame.simple_query("What is this?")
        men = last_frame.find("man")
        if len(men) == 0:
            men = [last_frame]
        man = men[0]
        man_action = man.simple_query("What is the man doing?")
        # Answer the question. Remember to create the info dictionary
        info = {
            "Caption of last frame": last_caption,
            "Man looks like he is doing": man_action
        }
        answer = video_segment.select_answer(info, question, possible_answers)
        return answer, info
    """


INSERT_IN_CONTEXT_EXAMPLES_HERE


Write a function using Python and the VideoSegment class (above) that could be executed to provide an answer to the query. 

Consider the following guidelines:
- Use base Python (comparison, sorting) for basic logical operations, left/right/up/down, math, etc.

INSERT_PREVIOUS_CODE_AND_ERROR_HERE

# INSERT_QUERY_HERE
\end{lstlisting}
\clearpage
\subsubsection{NExT-QA - Abstract API}
\begin{lstlisting}[language=Python]
import math


class VideoSegment:
    """A Python class containing a video, as well as relevant information.
    Attributes
    ----------
    video : np.ndarray
    A tensor of the original video.
    start : int
    An int describing the starting frame in this video segment with respect to the original video.
    end : int
    An int describing the ending frame in this video segment with respect to the original video.
    num_frames->int
    An int containing the number of frames in the video segment.

    Methods
    -------
    trim(start, end)->VideoSegment
    Returns a new VideoSegment containing a trimmed version of the original video at the [start, end] segment.
    select_answer(info, question, possible_answers)->str
    Returns the answer to the question given the possible answers and additional information.
    """

    def __init__(self, video: np.ndarray, start: int = None, end: int = None, parent_start=0, queues=None):
        """Initializes a VideoSegment object by trimming the video at the given [start, end] times and stores the
        start and end times as attributes. If no times are provided, the video is left unmodified, and the times are
        set to the beginning and end of the video.

        Parameters
        -------
        video : np.ndarray
        A tensor of the original video.
        start : int
        An int describing the starting frame in this video segment with respect to the original video.
        end : int
        An int describing the ending frame in this video segment with respect to the original video.
        """

        if start is None and end is None:
            self.trimmed_video = video
            self.start = 0
            self.end = video.shape[0] # duration
        else:
            self.trimmed_video = video[start:end]
        if start is None:
            start = 0
        if end is None:
            end = video.shape[0]
        self.start = start + parent_start
        self.end = end + parent_start

        self.num_frames = self.trimmed_video.shape[0]

    def trim(self, start: Union[int, None] = None, end: Union[int, None] = None) -> VideoSegment:
        """Returns a new VideoSegment containing a trimmed version of the original video at the [start, end]
        segment.

        Parameters
        ----------
        start : Union[int, None]
        An int describing the starting frame in this video segment with respect to the original video.
        end : Union[int, None]
        An int describing the ending frame in this video segment with respect to the original video.

        Examples
        --------
        >>> # Return the second half of the video
        >>> def execute_command(video):
        >>>     video_segment = VideoSegment(video)
        >>>     video_second_half = video_segment.trim(video_segment.num_frames // 2, video_segment.num_frames)
        >>>     return video_second_half
        """
        if start is not None:
            start = max(start, 0)
        if end is not None:
            end = min(end, self.num_frames)

        return VideoSegment(self.trimmed_video, start, end, self.start)

    def get_video_segment_of_event(self, event) -> VideoSegment:
        return get_video_segment_of_event(event)

    def get_video_segment_before_event(self, event) -> VideoSegment:
        return get_video_segment_before_event(event)

    def get_video_segment_after_event(self, event) -> VideoSegment:
        return get_video_segment_after_event(event)

    def caption_video(self, question) -> str:
        return caption_video(question)

    def simple_query(self, question) -> str:
        """Ask a simple question about the video.

        Examples
        --------
        # why does X happen?
        # possible_answers: ['answer1', 'answer2', 'answer3', 'answer4', 'answer5']
        def execute_command(video, question, possible_answers)->[str, dict]:
            # Create a video segment object
            video_segment = VideoSegment(video)
            # The question is simple, so just ask
            info = video_segment.simple_query("why does X happen?")
            # Choose the answer among given possible answers
            answer = select_answer(info, question, possible_answers)
            return answer
        """
        answer = simple_query(question)
        return answer


def select_answer(info: str, question: question, possible_answers: str) -> str:
    """Given an info, question and possible answers, select the correct answer.

    Examples
    --------
    # what does person A do after event X?
    # possible_answers: ['answer1', 'answer2', 'answer3', 'answer4', 'answer5']
    def execute_command(video, question, possible_answers)->[str, dict]:
        # Create a video segment object
        video_segment = VideoSegment(video)
        # Get video segment after event X
        video_segment_after = video_segment.get_video_segment_after_event("event X")
        # Ask what the person A is doing
        info = video_segment_after.caption_video("What is person A doing?")
        # Choose the answer among given possible answers
        answer = select_answer(info, question, possible_answers)
        return answer
    """


INSERT_IN_CONTEXT_EXAMPLES_HERE

Write a function using Python and the VideoSegment class (above) that could be executed to provide an answer to the query. 

Consider the following guidelines:
- Use base Python (comparison, sorting) for basic logical operations, left/right/up/down, math, etc.
- The input to your program is a video, question and possible answers.
- Always start your function by creating a `video_segment = VideoSegment(video)` object.

INSERT_PREVIOUS_CODE_AND_ERROR_HERE

# INSERT_QUERY_HERE
\end{lstlisting}

\end{document}